\pdfoutput=1

\documentclass[11pt]{article}

\usepackage[preprint]{acl}

\usepackage{color, xcolor}
\definecolor{lightgrey}{rgb}{0.83, 0.83, 0.83}
\definecolor{lightpurple}{rgb}{0.69, 0.61, 0.85}
\definecolor{lightblue}{rgb}{0.68, 0.85, 0.9}
\definecolor{lightorange}{rgb}{1.0, 0.8, 0.6}
\usepackage{fancyvrb} 
\usepackage{tcolorbox} 
\usepackage{graphicx}
\usepackage{tikz}

\usepackage{lipsum}

\usepackage{amssymb}
\usepackage{times}
\usepackage{latexsym}
\usepackage{amsmath}
\usepackage{threeparttable}
\usepackage{array}
\usepackage{makecell}
\usepackage{ulem}
\usepackage{colortbl}
\usepackage{natbib}
\usepackage{tcolorbox}

\usepackage[T1]{fontenc}

\usepackage[utf8]{inputenc}
\usepackage{booktabs}
\usepackage{comment}
\usepackage{microtype}
\usepackage{multirow}
\usepackage{inconsolata}

\usepackage{graphicx}

\title{Open-FinLLMs: Open Multimodal Large Language Models for Financial Applications}

\author{
\begin{tabular}{c}
Jimin Huang\textsuperscript{1}, Mengxi Xiao\textsuperscript{1}, Dong Li\textsuperscript{1}, Zihao Jiang\textsuperscript{1}, Yuzhe Yang\textsuperscript{4}, Yifei Zhang\textsuperscript{5}, \\
Lingfei Qian\textsuperscript{1}, Yan Wang\textsuperscript{1}, Xueqing Peng\textsuperscript{1}, Yang Ren\textsuperscript{1}, Ruoyu Xiang\textsuperscript{1}, Zhengyu Chen\textsuperscript{1}, \\
Xiao Zhang\textsuperscript{1}, Yueru He\textsuperscript{3}, Weiguang Han\textsuperscript{2}, Shunian Chen\textsuperscript{4}, Lihang Shen\textsuperscript{3}, Daniel Kim\textsuperscript{6}, \\
Yangyang Yu\textsuperscript{8}, Yupeng Cao\textsuperscript{8}, Zhiyang Deng\textsuperscript{8}, Haohang Li\textsuperscript{8}, Duanyu Feng\textsuperscript{9}, Yongfu Dai\textsuperscript{9}, \\
VijayaSai Somasundaram\textsuperscript{10}, Peng Lu\textsuperscript{11}, Guojun Xiong\textsuperscript{14}, Zhiwei Liu\textsuperscript{7}, Zheheng Luo\textsuperscript{7}, Zhiyuan Yao\textsuperscript{8}, \\
Ruey-Ling Weng\textsuperscript{1}, Meikang Qiu\textsuperscript{14}, Kaleb E Smith\textsuperscript{15}, Honghai Yu\textsuperscript{5}, Yanzhao Lai\textsuperscript{1}, Min Peng\textsuperscript{2}, \\
Jian-Yun Nie\textsuperscript{11}, Jordan W. Suchow\textsuperscript{8}, Xiao-Yang Liu\textsuperscript{3*}, Benyou Wang\textsuperscript{4*}, Alejandro Lopez-Lira\textsuperscript{10*}, \\
Qianqian Xie\textsuperscript{1,2*}, Sophia Ananiadou\textsuperscript{7,16,17}, Junichi Tsujii \textsuperscript{16}\\
\textsuperscript{1}The Fin AI \quad
\textsuperscript{2}Wuhan University \quad
\textsuperscript{3}Columbia University \quad\\
\textsuperscript{4}The Chinese University of Hong Kong, Shenzhen \quad
\textsuperscript{5}Nanjing University \quad \\
\textsuperscript{6}Rensselaer Polytechnic Institute \quad
\textsuperscript{7}The University of Manchester \quad \\
\textsuperscript{8}Stevens Institute of Technology \quad
\textsuperscript{9}National University of Singapore \quad \\
\textsuperscript{10}University of Florida \quad
\textsuperscript{11}University of Montreal \quad 
\textsuperscript{12}Yale University \quad
\textsuperscript{13}New York University \quad \\
\textsuperscript{14}Harvard University \quad
\textsuperscript{15}NVIDIA \\
\textsuperscript{16}Artificial Intelligence Research Centre \quad
\textsuperscript{17}Archimedes/Athena Research Centre \\
\texttt{xl2427@columbia.edu},  \texttt{wangbenyou@cuhk.edu.cn}, \\ \texttt{alejandro.lopez-lira@warrington.ufl.edu},  \texttt{xqq.sincere@gmail.com} \\
\textbf{* Corresponding Authors}
\end{tabular}
}

\begin{document}
\maketitle
\begin{abstract}
Financial LLMs hold promise for advancing financial tasks and domain-specific applications. However, they are limited by scarce corpora, weak multimodal capabilities, and narrow evaluations, making them less suited for real-world application.
To address this, we introduce \textit{Open-FinLLMs}, the first open-source multimodal financial LLMs designed to handle diverse tasks across text, tabular, time-series, and chart data, excelling in zero-shot, few-shot, and fine-tuning settings. The suite includes FinLLaMA, pre-trained on a comprehensive 52-billion-token corpus; FinLLaMA-Instruct, fine-tuned with 573K financial instructions; and FinLLaVA, enhanced with 1.43M multimodal tuning pairs for strong cross-modal reasoning.
We comprehensively evaluate Open-FinLLMs across 14 financial tasks, 30 datasets, and 4 multimodal tasks in zero-shot, few-shot, and supervised fine-tuning settings, introducing two new multimodal evaluation datasets. Our results show that Open-FinLLMs outperforms advanced financial and general LLMs such as GPT-4, across financial NLP, decision-making, and multi-modal tasks, highlighting their potential to tackle real-world challenges. 
To foster innovation and collaboration across academia and industry, we release all codes and models \footnote{\url{https://huggingface.co/collections/TheFinAI/open-finllms-66b671f2b4958a65e20decbe}} under OSI-approved licenses.
\end{abstract}

\section{Introduction}
The advancements of financial AI have been significantly driven by the progress of natural language processing (NLP) techniques, particularly large language models (LLMs) \citep{brown2020language,bubeck2023sparks}. Commercial LLMs like OpenAI's GPT-4 \citep{achiam2023gpt} and open-source LLMs such as Meta AI's LLaMA series \citep{touvron2023llama,touvron2023llama2openfoundation} have set new benchmarks in NLP tasks and vertical-domain tasks like medicine, owing to their impressive text understanding and generation capabilities. However, these general-purpose LLMs face limitations in the financial domain due to the knowledge gap. These models are primarily pretrained on general texts and lack an understanding of financial terminology, regulations, and market nuances \citep{xie2023wallstreetneophytezeroshot,wu2023bloomberggpt,xie2023pixiu}. Additionally, they are unable to effectively handle non-text data, such as tabular and time-series data, which are critical components of financial knowledge.

To bridge this gap, researchers have developed specialized financial LLMs through pre-training from scratch (BloombergGPT~\citep{wu2023bloomberggptlargelanguagemodel}), continual pre-training (FinTral~\citep{bhatia2024fintral}), or instruction tuning (PIXIU~\citep{xie2023pixiu}, FinGPT~\citep{liu2023fingpt,liu2024fingpt,yang2023fingpt}) using domain-specific data (Table~\ref{tab:comparison}). However, these models still face notable challenges~\citep{nie2024survey}:  
\textbf{First}, they rely on \textit{limited domain-specific corpora} for continual pre-training and instruction tuning, restricting their ability to fully capture the complexity of financial knowledge, language, and data types. For example, FinTral uses only 20 billion tokens for continual pre-training.  
\textbf{Second}, they show \textit{limited multimodal capabilities}, lacking support for tabular, time-series, and chart data. Most models, like BloombergGPT, focus solely on text, missing critical aspects of real-world tasks such as portfolio optimization and trend analysis.  
\textbf{Third}, evaluations are conducted on \textit{narrow scenarios}, mainly instruction-tuned financial NLP tasks. Zero/few-shot performance, multimodal reasoning, and financial decision-making tasks remain underexplored, limiting real-world applicability.

\begin{table*}[t]
\scriptsize
\centering
\renewcommand{\arraystretch}{1.3}
\caption{Comparison of key elements between Open-FinLLMs with other financial LLMs.
Abbreviations: PT for pre-training from scratch, CPT for continual pre-training, and IFT for instruction fine-tuning. 
}
\label{tab:comparison}
\scalebox{0.6}{
\begin{tabular}{@{}lccccccccccccc@{}}
\toprule
\multirow{2}{*}{\textbf{Model}} 
& \multirow{2}{*}{\textbf{Backbone}}
&\multirow{2}{*}{\textbf{Size}} 
& \multirow{2}{*}{\textbf{PT}}
&\multirow{2}{*}{\textbf{CPT}} 
& \multirow{2}{*}{\textbf{IFT}} 
& \multirow{2}{*}{\textbf{Tabular}} 
& \multirow{2}{*}{\textbf{Time}} 
& \multirow{2}{*}{\textbf{Chart}}
&\multicolumn{5}{c}{\textbf{Evaluation}}
\\ \cline{10-14}
 & & & & & & & & & \textbf{Zero-shot}
 &\textbf{Few-shot}
 &\textbf{Instruction-tuned}
 &\textbf{Multimodal}
 &\textbf{Trading}
\\ 
\midrule
\textbf{BloombergGPT}~\citep{wu2023bloomberggptlargelanguagemodel}  & BLOOM & 50B & 363B&$\times$ & $\times$ & $\times$ & $\times$&$\times$&$\times$&\checkmark&$\times$&$\times$&$\times$\\ 
\midrule
\textbf{PIXIU}~\citep{xie2023pixiu} & LLaMA & 7/30B &$\times$ &$\times$ & 128K & $\times$ & $\times$&$\times$&$\times$&$\times$&\checkmark&$\times$&$\times$\\ \midrule
\textbf{FinGPT}~\citep{liu2023fingpt} & LLaMA2 & 7B &$\times$ &$\times$ & 205.3K & $\times$ & $\times$&$\times$&$\times$&$\times$&\checkmark&$\times$&$\times$\\ \midrule
\textbf{FinTral}~\citep{bhatia2024fintral} &Mistral&
7B & $\times$&20B & 226.3K&$\times$&$\times$&\checkmark&$\times$&$\times$&\checkmark&\checkmark&$\times$\\\midrule
\textbf{Open-FinLLMs} & LLaMA3&8B&$\times$&52B&573K&\checkmark&\checkmark&\checkmark&\checkmark&\checkmark&\checkmark&\checkmark&\checkmark\\ \bottomrule
\end{tabular}}
\end{table*}

To address these limitations, in this paper, we introduce \textit{Open-FinLLMs}, a series of financial large language models tailored for various financial tasks. We begin with FinLLaMA, a groundbreaking foundational model pre-trained on a massive \textbf{52-billion-token} corpus comprising text, tabular, and time-series data from high-quality financial sources such as reports, papers, and market data for the first time.
This innovative pre-training strategy incorporating extensive data modalities equips FinLLaMA with deep financial insights and analytical capabilities.
We further develop FinLLaMA-Instruct by fine-tuning the
model with expanded datasets of \textbf{573K} diverse and high-quality financial instructions. Enriched instruction-tuning datasets enhance not only the model's ability to follow instructions but also coverage of financial domain knowledge, leading to better performance for a wide range of downstream tasks.
To handle multimodal financial data, we present FinLLaVA, leveraging \textbf{1,430K} financial multimodal instruction pairs, including images, text, charts, and tabular data. Unlike traditional methods that primarily focus on standard image-text pairs, our fine-tuning process is the first to incorporate chart image-text pairs and images of tabular layouts. This innovation enables the model to effectively interpret and process complex financial data with improved precision and versatility.

We comprehensively evaluate our models across 14 financial tasks and 30 datasets. FinLLaMA is tested on 19 datasets (9 tasks) in zero-shot and 4 datasets (3 tasks) in few-shot settings.
It outperforms other financial and general LLMs including LLaMA3-8B, LLaMA3.1-8B, and BloombergGPT in almost all tasks (including broad financial NLP, reasoning, asset trading and decision making tasks) in the zero-shot and few-shot settings, showing strong generalization .
FinLLaMA-Instruct outperforms other financial LLMs on 4 out of 6 domain-specific tasks like sentiment analysis, NER, numeric understanding, summarization, stock prediction, and credit scoring, also surpassing GPT-4 on 3 tasks.
While FinLLaVA outperforms other open multimodal LLMs on 4 multimodal tasks (1 general, 3 financial), and even surpasses advanced close source models including GPT-4o and Gemini-1.5-pro on the tabular task,  Together, these highlight Open-FinLLMs as powerful tools for real-world financial applications.
In summary, our key contributions are\footnote{We will publicly share the training code, datasets, and models.}:
(1) We introduce Open-FinLLMs, a groundbreaking suite of financial LLMs trained on three comprehensive datasets tailored for different training stages. This structured training pipeline ensures a robust understanding of financial terminology, numerical data, and complex financial contexts. 
(2) By pioneering the integration of tabular and time-series data, Open-FinLLMs are the first financial foundation models with advanced multimodal capabilities. This multimodal approach bridges the gap between textual and structured financial data, enabling a more comprehensive capture of financial knowledge from diverse data types.
(3) Extensive experiments demonstrate Open-FinLLMs' superior performance across a range of financial and multimodal benchmark tasks, including zero-shot and few-shot settings. Notably, the models excel in both traditional NLP tasks and finicial tasks in real applications, showcasing their readiness to tackle complex challenges understanding multimodal finicial data.
(4) We have open-sourced the training code, datasets, and models, providing a valuable reference for future research on LLMs in the financial domain. The shared code and training methodologies can also support advancements in other fields that require processing time-series data or understanding multimodal information.

\section{Related Work}
\subsection{Financial Large Language Models} 
Recently, several financial LLMs have been developed to address domain-specific challenges. BloombergGPT~\citep{wu2023bloomberggpt} pioneered financial LLMs with pretraining on 363 billion financial tokens; however, it remains closed-source, limiting its accessibility. On the open-source front, PIXIU \citep{xie2023pixiu} and FinGPT \citep{liu2023fingpt,liu2024fingpt,yang2023fingpt} fine-tuned LLaMA models with hundreds of thousands of financial instructions, focusing primarily on text-based tasks. FinTral \citep{bhatia2024fintral} introduced multimodal capabilities for tabular data, but limited by the small size of its domain-specific data and its evaluation scope. Fin-o1 \cite{qian2025fino1}, developed through instruction tuning on reasoning-enhanced data, demonstrates that domain-enhanced reasoning abilities significantly improve the model's performance in reasoning tasks.

\subsection{Domain Specialization of Large Language Models}
Domain-specific LLMs are developed using three main strategies: pre-training from scratch (PT), continued pre-training (CPT), and instruction fine-tuning (SFT)~\citep{wu2024continual}.
CPT adapts existing LLMs by further training on domain data, like Code LLaMA~\citep{roziere2023code}, which improves code generation. SFT tailors models for specific tasks with domain instructions, requiring less data and compute, as demonstrated by PIXIU~\citep{xie2023pixiu} and FinGPT~\citep{liu2024fingpt}.

\section{Open-FinLLMs: Open Multimodal Financial LLMs}
\begin{figure*}[t]
\centering
\includegraphics[width=0.9\linewidth]{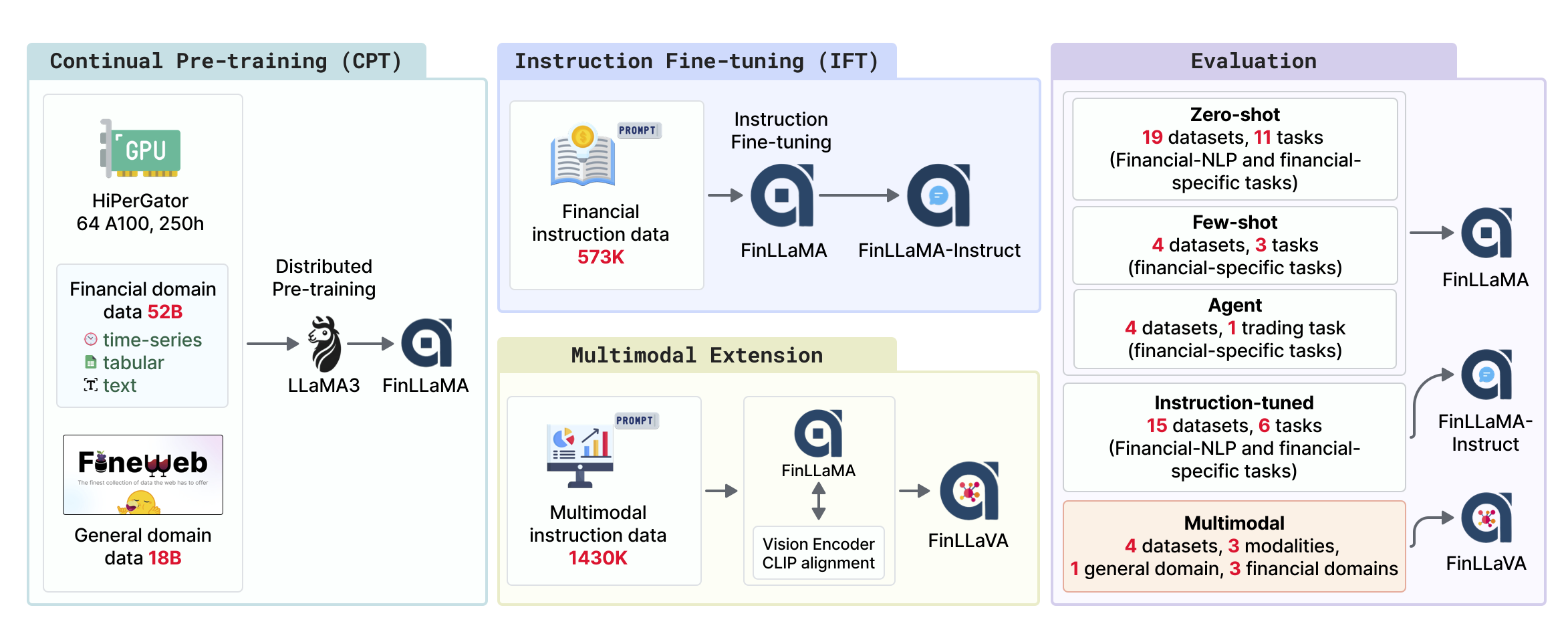}
\caption{Overview of Open-FinLLMs.}
\label{fig:overview}
\end{figure*}

In this section, we introduce the Open-FinLLMs model family as shown in Figure \ref{fig:overview}, including FinLLaMA for foundational financial knowledge, FinLLaMA-Instruct for instruction-following tasks, and FinLLaVA for multimodal financial applications.

\subsection{FinLLaMA: Specializing LLaMA3 for Finance with Continual Pre-training}
\textbf{Curation of Continual Pre-training Corpus.}
To facilitate effective continual pretraining, we construct a comprehensive financial corpus comprising 52 billion tokens sourced from seven diverse financial domains.
\begin{table}[t]
\scriptsize
\centering
\renewcommand{\arraystretch}{1.3}
\caption{Comparison of financial corpora used for pre-training BloombergGPT~\citep{wu2023bloomberggptlargelanguagemodel}, FinTral~\citep{bhatia2024fintral}, and Open-FinLLMs.}
\label{tab:pretrain-data-comparison}
\resizebox{\columnwidth}{!}{%
\begin{tabular}{lccc}
\hline
\textbf{Data Types} & \textbf{BloombergGPT} & \textbf{FinTral} & \textbf{Open-FinLLMs} \\ \hline
Financial Papers & - & - & \textbf{4B} \\
Conference Calls & - & - & \textbf{5B} \\
Financial Reports & \textbf{9B} & 1.55M & 5B \\
Indicators & - & - & \textbf{12B} \\
News+Social Media & \textbf{43B} & 5.657B & 7B \\
Historical Data & - & - & \textbf{13B} \\
SEC Filings & \textbf{14B} & 2.55B & 6B \\
Web Data & \textbf{298B} & 11.75B & - \\
\hline
Total & \textbf{363B} & 20.0B & 52B \\
\hline
\end{tabular}}
\end{table}
As shown in Table \ref{tab:pretrain-data-comparison}, our continual pre-training corpus encompasses a wide array of data sources to ensure comprehensive coverage of financial knowledge: (1) \textit{Financial papers}: $4$ billion tokens from academic papers and research articles, offering a strong foundation in financial concepts and theories. These papers, sourced from SSRN\footnote{https://www.ssrn.com/index.cfm/en/} and open-source conference proceedings, span from 2000 to 2023 and cover topics such as market analysis, financial modeling, and economic theory.
(2) \textit{Conference calls}: 5 billion tokens of open-source transcripts from earnings calls, analyst meetings, and investor briefings, collected from 09/08/2004 to 12/17/2021, providing real-time insights into corporate performance and strategies.
(3) \textit{Financial reports}: This component consists of 5 billion tokens from annual and quarterly reports, covering the period from 2005 to 2020, crucial for assessing a company's financial health and market positioning.
(4) \textit{Technical indicators}: 12 billion tokens of economic indicators and financial ratios sourced from company filings and open market data spanning from 2009 to 2023, essential for macroeconomic analysis and investment decisions.
(5) \textit{News and social media}: 7 billion tokens from financial news outlets and social media platforms collected from 1999 to 2021, offering timely updates on market trends and public opinion.
(6) \textit{Historical data}: 13 billion tokens of historical stock prices, trading volumes, and market data from 1999 to 2022, vital for quantitative analysis and algorithmic trading.
(7) \textit{SEC filings}: 6 billion tokens from U.S. SEC filings, such as 10-K and 10-Q reports spanning from 1994 to 202, providing comprehensive insights into corporate activities and performance.
Unlike existing models, we chose not to use web data due to its higher noise level compared to other data sources. See Appendix \ref{data-building} for more details of the pre-training data.

\textbf{Mixing Financial Data with General Data}.
To prevent catastrophic forgetting, as highlighted in previous studies \citep{wu2023bloomberggpt,guptacontinual}, we integrate a subset of general-domain data with our financial-domain corpus. We use the Fineweb dataset \citep{penedo2024fineweb}, which contains over 15 trillion tokens of cleaned and deduplicated English web data from CommonCrawl. Using the approach outlined in DoReMi \citep{xie2024doremi}, we determine the optimal mixture ratio of financial to general-domain data to be around 3:1. Accordingly, we sampled a total of 18 billion tokens from the general-domain corpus. This mixture helps our models retain previously learned general knowledge while being specifically fine-tuned for financial tasks.

\textbf{Pretraining Details.}
We use LLaMA3-8B as the backbone, optimizing the token sequence likelihood with a standard language modeling objective. The model maximizes $\mathcal{L}(\Theta) = \sum_{i}^k \log P_{\Theta}(x_i|x_1,\dots,x_{i-1})$ to predict token sequences. Distributed training is performed using DeepSpeed on 64 A100 80GB GPUs across 8 nodes, requiring ~250 hours for 1 epoch. The learning rate is set to $1 \times 10^{-5}$ with a cosine schedule, weight decay of 0.00001, and a 0.05 warm-up ratio. The batch size is 2 per device, with a maximum sequence length of 8,192 tokens.

\subsection{FinLLaMA-Instruct: Domain Task Optimization through Instruction Tuning}
Building on the foundation of FinLLaMA, we propose FinLLaMA-Instruct-8B, developed through instruction fine-tuning to enhance the model's instruction-following capabilities and optimize performance on downstream domain tasks.

\textbf{Financial Instruction Dataset}
To optimize FinLLaMA for downstream domain tasks and instruction following ability, we have assembled an extensive instruction-tuning dataset, totaling 573K samples, specifically tailored for financial applications, as shown in Table~\ref{tab:instruction_datasets}. 
The dataset is sourced from four open data sources: (1) FLUPE~\citep{xie2023pixiu}, with 123K samples covering key financial NLP tasks; (2) finred~\citep{liu2023fingpt}, with approximately 32.67K examples focused on financial report and document comprehension; (3) MathInstruct~\citep{yue2023mammoth}, with 262K examples from 13 distinct mathematical rationale datasets; and (4) Sujet-Finance-Instruct-177k, which integrates data from 18 different financial NLP datasets.
To ensure the uniqueness and quality of the data, we performed rigorous deduplication and filtering, addressing potential overlaps between datasets like FLUPE and Sujet-Finance-Instruct-177k. See Appendix \ref{instruction-data} for more details on the instruction tuning data and data processing.
\begin{table}[ht]
\centering
\caption{Overview of instruction datasets used in FinLLaMA-Instruct and comparison with FinTral's data.}
\scriptsize 
\label{tab:instruction_datasets}
    \begin{tabular}{lrc}
    \toprule
    \textbf{Source} & \textbf{Fintral} & \textbf{FinLLaMA-Instruct} \\ 
    \midrule
    ChanceFocus/FLUPE & 123.0k & 123.0k \\ 
    FinGPT/Hingpt-finred & 32.67k & 32.67k \\ 
    TIGER-Lab/MathInstruct & 26.2k & 262k \\ 
    sujet-ai/Sujet-Finance-Instruct-177k & - & 177k \\ 
    \midrule
    \textbf{Total after deduplication} & 226.3k & \textbf{573k} \\
    \bottomrule
    \end{tabular}
\end{table}

\textbf{Instrution-tuning Details.}
For instruction tuning, we utilize FinLLaMA as the backbone model and conduct training on 8 NVIDIA A100 80GB GPUs for 6 hours. The model is optimized using Qlora~\citep{dettmers2024qlora} via AutoTrain~\footnote{https://huggingface.co/autotrain}, configured with a block size and model maximum length of 4096. We train the model over 2 epochs with a batch size of 1 and a learning rate of 0.0002. Parameter-efficient tuning is achieved with LoRA settings of \( r = 64 \), \(\alpha = 128\), and no dropout, using INT4 quantization. All linear modules are targeted, with right-aligned padding. The AdamW optimizer~\citep{loshchilovdecoupled}, coupled with a cosine scheduler, is used for optimization, along with gradient accumulation set to 4.

\subsection{FinLLaVA: Enabling Multimodal Capabilities via Multimodal Instruction Tuning}
Based on FinLLaMA, we build its financial multi-modal extension FinLLaVA to address multimodal financial tasks by leveraging multimodal instruction tuning based on the LLaVA-1.5~\citep{liu2023improved} framework. 
\begin{table}[t]
\scriptsize
\centering
\caption{Statistics of multimodal instruction dataset. SFT stands for supervised fine-tuning. Asterisk (*) indicates the dataset only contains textual data.}
\scriptsize
\label{tab:visual-data}
\resizebox{\columnwidth}{!}{
\begin{tabular}{cccc}
\hline
\textbf{Stage} 
& \textbf{Dataset}  
& \textbf{Source} 
& \textbf{Instructions} 
\\ \hline
\multirow{6}{*}{\textbf{\shortstack{Multimodal\\Alignment}}}
& ALLaVA-4V
& \citep{chen2024allava}
& 468k
\\
& OCR-VQA
& \citep{wang2023believepromptinggpt4vbetter}
& 79k
\\
& SynthTabNet
& \citep{nassar2022tableformer}
& 20k
\\
& UniChart
& \citep{masry2023unichart}
& 5k
\\
& ChartQA         
& \citep{masry-etal-2022-chartqa}
& 20k 
\\
& Chart2Text         
& \citep{obeid2020charttotextgeneratingnaturallanguage}
& 30k
\\ \hline
\multirow{2}{*}{\textbf{SFT}}
& LLaVA-v1.5-mix665k
& \citep{liu2023llava}
& 665k
\\
& Evol-Instruct *
& \citep{chen2024allava}
& 143k
\\ \hline
\textbf{Total}
& 
& 
& 1430k
\\ \hline
\end{tabular}%
}
\end{table}

\textbf{Multimodal Instruction Data.}
We curated a diverse multi-modal dataset comprising image, tabular, chart, and text data to ensure comprehensive coverage of various data formats, as shown in Table \ref{tab:visual-data}.
For image data, we utilized three vision instruction datasets: ALLaVA-4V~\citep{chen2024allava}, LLaVA-v1.5-mix665k~\citep{liu2023llava}, and OCR-VQA~\citep{wang2023believepromptinggpt4vbetter}.
For chart data, we integrate subsets from multiple sources: UniChart \citep{masry2023unichart}, with 5k chart image-text pairs; Chart2Text \citep{obeid2020charttotextgeneratingnaturallanguage}, with 30k chart image-text pairs; and ChartQA \citep{masry-etal-2022-chartqa}, featuring 20k chart images and their associated question-answering (QA) pairs.
We used GPT-4o to evaluate these datasets and filtered out images most relevant to the financial domain.
Further details are provided in Appendix~\ref{chart}.
Different from previous work~\citep{bhatia2024fintral}, which only image and chart data, we selected a subset of the SynthTabNet~\citep{nassar2022tableformer} dataset, consisting of 20k annotated images of data in tabular layouts.
Further details are provided in Appendix~\ref{table}. Additionally, we included Evol-Instruct \citep{chen2024allava}, a dataset of 143k pure text instructions, to enhance the model’s generalization capabilities and reduce the risk of hallucination.


\textbf{Multimodal Instruction Finetuning.}
We utilize CLIP \citep{radford2021learningtransferablevisualmodels} as our visual encoder in conjunction with the FinLlaMA language decoder, fine-tuning the model on our multimodal instruction dataset. Our approach follows the training framework established by LLaVA-1.5 \citep{liu2023llava}, implementing a two-stage instruction-tuning process.
\textbf{Multimodal Alignment:}
In this initial stage, we aim to align the vision encoder's output with the language model's embedding space. During this phase, both the vision encoder and LLM weights remain frozen. The key objective is to train a two-layer MLP projector to bridge the gap between the vision encoder's features and the LLM's embedding. For each input, consisting of an image $X_v$, instructions $X_{\text{instruct}}$ that may involve single-turn or multi-turn conversations, and the target answer $X_a$, the vision encoder processes the image data to generate a vision feature: $Z_v = g(X_v)$. The MLP projector then maps $Z_v$ into the embedding space of the language model: $H_v = f_{\text{MLP}}(Z_v;\theta)$, where $\theta$ represents the trainable parameters of the projector. The training objective is to maximize the auto-regressive likelihood:
$\sum_i^L P_\theta(X_i \mid X_v, X_{\text{instruct}, <i}, X_{a, <i})$, where $L$ is the sequence length of the target answer $X_a$, and $X_{\text{instruct}, <i}$ and $X_{a, <i}$ are the tokens of instructions and answers preceding the current prediction $X_i$.
\textbf{Supervised Fine-tuning:}
In the second stage, we continue updating the parameters of both the language model and the MLP projector, while keeping the vision encoder's parameters frozen. We maintain the same autoregressive training objective as the previous stage but apply it to a different dataset. As shown in Table \ref{tab:visual-data}, we utilize LLaVA-v1.5-mix665k and Evol-Instruct as our primary training data sources.


\textbf{Training Details:} 
In the multi-modal alignment stage, we set the global batch size to 128, the learning rate to $1 \times 10^{-3}$, with a warm-up ratio of 0.03 and cosine decay. Training uses \texttt{bf16} and \texttt{tf32} precision for stability and acceleration. Weight decay is set to 0.0, and the model's maximum length is 2048 tokens. We train on eight NVIDIA HGX H20 GPUs, completing one epoch in approximately 30 hours.
In the SFT stage, we set the global batch size to 256, the learning rate to $2 \times 10^{-5}$, with a warm-up ratio of 0.05 and cosine decay. The model's maximum length is increased to 8192 tokens for longer sequences. Weight decay remains at 0.0, and training runs for one epoch with the same precision settings for efficiency and performance.

\section{Experiments}
We conducted broad evaluations on general and financial tasks. Unlike FinTral \cite{bhatia2024fintral}, which only reported instruction-tuned performance, we assess: 1) zero- and few-shot performance of the FinLLaMA base model, 2) FinLLaMA-Instruct performance, 3) trading performance, and 4) multimodal capabilities.

\subsection{Performance of FinLLaMA}
In this section, we evaluate the continual pretrained FinLLaMA model using 26 datasets spanning 11 critical financial tasks, categorized into financial-NLP tasks (Sentiment Analysis, Classification, QA, and Named Entity Recognition) and financial-specific tasks (Misinformation, Math, Credit Scoring, Fraud Detection, Financial Distress, Claim Analysis,  Decision Making, Auditing and Risk \& Compliance). We compare its performance against competitor models by reproducing their results where possible or reporting the best publicly available scores.

\begin{table*}[!htbp]
\centering
\scriptsize
\caption{Performance of FinLLaMA and baseline models on benchmark tasks, ranging from 0 to 100. We boldface the best performance in each benchmark task. (-) indicates N/A.}
\label{tab:mffm_results}
\resizebox{0.7\textwidth}{!}{
    \begin{tabular}{@{}lcccccc@{}}
    \toprule
    \textbf{Category} 
    & \textbf{Task} 
    & \textbf{\makecell[c]{Dataset}}
    & \textbf{\makecell[c]{LLaMA3-8B}}
    & \textbf{\makecell[c]{LLaMA3.1-8B}} 
    & \textbf{\makecell[c]{BloombergGPT}}
    & \textbf{\makecell[c]{FinLLaMA}} \\ 
    \midrule
    \textbf{Financial} & Sentiment Analysis & TSA & 75.00 & 67.00 & - &  \textbf{81.00} \\ 
    (zero-shot) & Classification & FOMC & 41.00 & 47.00 & - & \textbf{50.00} \\ 
    & Classification & FinArg-AUC & 51.00 & 51.00 & - & \textbf{55.00} \\ 
    & Classification & MA & 34.00 & 51.00 & - & \textbf{70.00} \\
    & Classification & SC & 69.00 & 73.00 & - & \textbf{86.00} \\ 
     & Misinformation & FinFact & 29.48 & 32.07 & - & \textbf{34.62} \\ 
     & Math & MC  & 15.00 & \textbf{19.3} & - & {18.00} \\ 
     & Math & KnowledgeMath & 2.3 & \textbf{2.7} & - & 2.5 \\
     & Math & DocMath-Eval & 1.8 & \textbf{3.5} & - & 3.1 \\
     & Credit Scoring & German  & 34.00 & \textbf{66.02} & -  & 34.00 \\
     & Credit Scoring & Australian & 27.60 & 26.00 & - & \textbf{49.80} \\
      & Credit Scoring & LendingClub & 9.30 & \textbf{38.00} & - & 22.60 \\ 
     & Fraud Detection  & ccf  & 50.10 & 50.06 & -& \textbf{50.10} \\ 
      & Fraud Detection  & ccfraud  & 49.20 & 48.02 & - & \textbf{50.25} \\
      & Financial Distress  & polish & 47.65 & 50.00 & - & \textbf{50.00} \\
       & Financial Distress & taiwan & 45.80 & 47.75 & - & \textbf{50.00} \\
       & Claim Analysis & ProtoSeguro & 48.95 & 49.35 & - & \textbf{49.55} \\
       & Claim Analysis & travelinsurance & 50.00 & 50.00 & - & \textbf{50.55} \\
       & QA & ConvFinQA & 31.95 & 32.55 & 43.41 & \textbf{51.41}\\
       & Risk\& Compliance & RegulationQA & 16.72 & 17.35&&\textbf{17.34}\\
       & Auditing  & Abbreviation & 14.68 & 15.99&&\textbf{18.92}\\
    \midrule
    (5-shots) & Sentiment Analysis & FPB & 69.65 & 13.08 & 51.07 & \textbf{70.25} \\ 
     & Sentiment Analysis & FiQA-SA & 52.29 & 65.39 & 75.05 & \textbf{75.34} \\ 
    & Classification & Headlines & 80.59 & 59.95 & 82.20 & \textbf{85.54} \\ 
    \midrule
    (20-shots) & Named Entity Recognition & NER & 39.18 & 49.04 & 60.82 & \textbf{82.10}  \\
    \midrule
    Agent & Decision Making & Single-asset trading & 45.50 & 45.50 & 43.50 & \textbf{67.67} \\ 
    \bottomrule
    \end{tabular}
}
\end{table*}

\textbf{Zero-shot Performance}
In zero-shot scenarios, as shown in Table \ref{tab:zero-shot-dataset-info}, our evaluation uses 19 datasets covering both financial-NLP and financial-specific tasks,
A detailed description of evaluation tasks and detailed prompts for each dataset are available in Appendix \ref{append:basemodel_eval_task_details} and Appendix \ref{append:icl_prompt}.

Overall, Table \ref{tab:mffm_results} (and Figure \ref{fig:zero-shot-performance} in Appendix \ref{appendix:zero-shot}) demonstrates that FinLLaMA outperformed the baseline models on most financial tasks, highlighting its robustness and versatility in zero-shot settings.
It surpasses its backbone model, LLaMA3-8B, on all tasks, highlighting the effectiveness of continual pre-training with large-scale domain-specific data in enhancing financial knowledge. Additionally, FinLLaMA exceeds the performance of BloombergGPT, despite its larger model size of 50B, and also outperforms the current most capable open-source LLM, LLaMA3.1-8B on most tasks. 

FinLLaMA shows exceptional performance in sentiment analysis and classification tasks, demonstrating its proficiency in fundamental financial operations. Its improved accuracy in fact-checking, as shown by the FinFact dataset, highlights its ability to comprehend and evaluate financial information, enabling precise judgments on claims. In math problem-solving tasks, our model shows improvement across all datasets (MC, KnowledgeMath, and DocMath-Eval) compared to LLaMA3-8B.

In credit scoring tasks, performance varies significantly across datasets. Upon manual inspection of prediction results, we found that on the German dataset, both our model and LLaMA3-8B predicted all cases as one category, while LLaMA3.1-8B predicted the opposite. On the Australian dataset, our model's superior performance demonstrates the benefits of continuous pre-training, even with anonymized features. In the LendingClub dataset, our model outperformed LLaMA3-8B, though not LLaMA3.1-8B, likely due to its larger scale training data. Additionally, FinLLaMA excelled in fraud detection, financial distress identification, and claim analysis, showcasing its robust capabilities across diverse financial tasks.

\textbf{Few-shot Performance} 
For the few-shot evaluation setting, we use four datasets covering three financial-NLP tasks, as shown in Table \ref{tab:few-shots-dataset-info}. These tasks are aligned with the BloombergGPT evaluation settings to ensure consistency and comparability. Detailed descriptions of the evaluation tasks can be found in Appendix \ref{append:few_shot_task_desrip}.

Overall, as shown in Fig. \ref{fig:few-shot-performance}, FinLLaMA consistently outperforms baseline models across a wide range of financial tasks, demonstrating its robustness and versatility in few-shot settings.
In the NER task, FinLLaMA achieves a remarkable F1 score of 82.10, significantly surpassing its backbone model LLaMA3-8B (39.18), LLaMA3.1-8B (49.04), and BloombergGPT (60.82). This highlights the substantial improvement in entity recognition due to continual pre-training.
For the FPB dataset, FinLLaMA achieves an F1 score of 70.25 in few-shot settings, outperforming LLaMA3-8B (69.65), LLaMA3.1-8B (13.08), and BloombergGPT (51.07). Similarly, in the FiQA-SA dataset, FinLLaMA scored 75.34, surpassing LLaMA3-8B (52.29), LLaMA3.1-8B (65.39), and BloombergGPT (75.05).
In the Headlines dataset for classification, FinLLaMA achieves a score of 85.54, outperforming LLaMA3-8B (80.59), LLaMA3.1-8B (59.95), and BloombergGPT (82.20). These results demonstrate FinLLaMA's strong ability to classify financial texts accurately with minimal examples.

\textbf{Agent}
We further assess the performance of FinLLaMA on a financial-specific task (Decision-Making) using the FinMem agent framework \citep{yu2024finmem}, tested on a single-asset trading dataset comprising multi-sourced financial data from August 15, 2021, to April 25, 2023. The results are presented in Table \ref{tab:trading-performance}. This task assesses the LLM's proficiency in single-asset trading, with Cumulative Return and Sharpe Ratio as the key performance metrics. For more tasks and datasets, please refer to Appendix~\ref{trading-details}.
As shown in Table~\ref{tab:trading-performance}, FinLLaMA outperforms other LLMs with positive Cumulative Return and Sharpe Ratio metrics, demonstrating profitability in dynamic trading environments. It achieves the highest Sharpe Ratio (SR) of over 1, indicating a superior risk-return balance. Additionally, FinLLaMA maintains investment stability with an Annual Volatility of 0.4766 and a Maximum Drawdown of 0.2693, both lower than other models. This combination of high Sharpe Ratio and low volatility highlights FinLLaMA's ability to deliver consistent returns with minimized risk, making it highly reliable for trading strategies. These results highlight the significant impact of continual pre-training in enhancing FinLLaMA's performance. For detailed results, please refer to Appendix~\ref{trading-comparison}.

\subsection{Performance of FinLLaMA-Instruct}
Table \ref{inst_eval_datasets} provides detailed information on the datasets and tasks used for the evaluation of instruction fine-tuned models.
We align our evaluations with Fintral's~\citep{bhatia2024fintral} settings for consistency and comparability, reporting the average performance across datasets in each task. The evaluation includes 6 tasks and 15 datasets used in the FinBEN paper~\citep{xie2024finben}:
(1) sentiment analysis (SA) task, using FiQA-SA, FOMC, FPB, and Headlines datasets; (2) named entity recognition (NER) task, using Finer-Ord and NER datasets; (3) number understanding (NU) task, using ConvFinQA and FinQA datasets; (4) text summarization (TS) task, using ECTSUM and EDTSUM datasets; (5) stock movement prediction (SMP) task, using ACL18, BigData22, and CIKM18 datasets; (6) credit scoring (CS) task, using Australia and German datasets. 

Overall, Table \ref{table:inst_performance} shows that FinLLaMA-Instruct outperforms other financial LLMs on 4 out of 6 financial tasks, including surpassing GPT-4 on 3 tasks, underscoring its effectiveness and applicability in the financial domain.
FinLLaMA-Instruct also exceeds the performance of all other specialized financial LLMs, including Palmyra-Fin-70B-32, which is significantly larger, in 4 out of 6 financial tasks.
In the numerical understanding task, FinLLaMA-Instruct achieves the best performance with an average accuracy score of 0.69, even outperforming GPT-4. This highlights the effectiveness of our instruction tuning using large-scale math reasoning data in enhancing the model's numeric understanding ability.
Furthermore, FinLLaMA-Instruct consistently outperforms Mistral-7B-Instruct on all tasks and surpasses ChatGPT in five out of six tasks, with text summarization being the exception. This underscores its superior performance compared to general LLMs.
FinLLaMA-Instruct achieves better performance compared with GPT-4 on three key financial analysis tasks, demonstrating the robustness of the FinLLaMA backbone model and the effectiveness of our fine-tuning approach and datasets.

\begin{table}[!h]
\scriptsize
\centering
\caption{Performance of FinLLaVA and baseline models on zero-shot multi-modal benchmark evaluations. Asterisks (*) indicate results on the MMMU test dataset \citep{yue2023mmmu}. Daggers (\textsuperscript{\dag}) indicate results on our own benchmarks.}
\label{tab:multimodal_results}
\resizebox{\columnwidth}{!}{%
\begin{tabular}{m{0.95cm}ccm{1.5cm}m{1.5cm}m{1.8cm}m{1.8cm}c}
\hline
\textbf{} 
& \textbf{Method}  
& \textbf{Backbone} 
& \textbf{MMMU-Overall*}
& \textbf{MMMU-Bussniess*} 
& \textbf{ChartBench\textsuperscript{\dag}} 
& \textbf{TableBench\textsuperscript{\dag}}
\\ \hline
\multirow{4}{*}{\textbf{\shortstack{Closed\\Source}}}
& Gemini-1.5-pro
& -
& 49.3
& 49.8
& 61.4
& 58.2
\\
& GPT-4o
& -
& \textbf{55.7}
& \textbf{64.3} 
& \textbf{66.3}
& 66.7
\\
& Qwen-VL-MAX
& -
& 46.8
& 39.8
& 56.0
& 55.4
\\ \hline
\multirow{7}{*}{\textbf{\shortstack{Open\\Source}}}
& LLaVA-1.5
& Vicuna-7B
& 32.0
& 26.3
& 43.4
& 56.0
\\
& LLaVA-1.5
& Vicuna-13B
& 33.6
& 29.0
& 49.1
& 69.1
\\
& LLaVA-1.6
& Vicuna-7B
& 32.3
& 25.6
& 41.7
& 42.9
\\
& LLaVA-1.6
& Vicuna-13B
& 34.0
& 29.1
& 50.3
& 59.3
\\
& Deepseek-VL-7B-Chat
& DeepSeek-LLM-7B-Base
& 34.2
& 28.6
& 51.4
& 57.3
\\
& Qwen-VL-Chat
& Qwen-7B
& 32.0
& 26.2
& 52.6
& 48.2
\\
\rowcolor{blue!20}
& FinLLaVA
& FinLLaMA
& 36.3
& 30.7
& 52.9  
& \textbf{72.4}
\\ \hline
\end{tabular}}
\end{table}

\begin{table}[h]
\scriptsize
\renewcommand{\arraystretch}{1}
\caption{Performance of FinLLaMA-Instruct and baselines on benchmark tasks (0-1 range). Best performances are bolded, and second-best are underlined. FinTral results are cited from its paper due to unavailable evaluation code or methodology.} 
\label{table:inst_performance}
\begin{threeparttable}
\resizebox{\columnwidth}{!}{
\begin{tabular}{lcccccc}
\hline
\textbf{Model}
& \textbf{SA}
& \textbf{NER}   
& \textbf{NU}     
& \textbf{TS}      
& \textbf{SMP}    
& \textbf{CS}      
\\ \hline
Mistral-7B-Instruct-v0.1\tnote{1}~\citep{jiang2023mistral}
& 0.49
& 0.00
& 0.00 
& 0.30 
& 0.49 
& 0.48
\\
ChatGPT (gpt-3.5-turbo) 
& 0.70
& 0.53 
& 0.58
& \uline{0.59}
& 0.53 
& 0.31
\\
GPT-4 (gpt-4-0613)~\citep{openai2023gpt4}
& 0.79
& \textbf{0.80} 
& \uline{0.63}
& \textbf{0.65}
& \uline{0.54} 
& \textbf{0.70}
\\
Fintral~\citep{bhatia2024fintral}     
& \uline{0.81}       
& 0.40     
& 0.02     
& {0.40} 
& 0.53    
& \uline{0.61} 
\\
Palmyra-Fin-70B-32K\tnote{2}
& 0.69      
& 0.08    
& 0.21  
& 0.07      
& 0.54      
& 0.53      
\\ 
FinMA-7B-full\tnote{3}~\citep{xie2023pixiu}
& 0.78 
& 0.35 
& 0.12
& 0.35 
& 0.51 
& 0.29
\\ \hline
FinLLaMA-instruct       
& \textbf{0.82} 
& \uline{0.57}
& \textbf{0.69} 
& 0.37       
& \textbf{0.56} 
& 0.56     
\\ \hline
\end{tabular}}
\begin{tablenotes}
\scriptsize
\item[1] https://huggingface.co/mistralai/Mistral-7B-Instruct-v0.1
\item[2] https://huggingface.co/Writer/Palmyra-Fin-70B-32K
\item[3] https://huggingface.co/TheFinAI/finma-7b-full 
\end{tablenotes}
\end{threeparttable}
\end{table}

\subsection{Performance of FinLLaVA}
\textbf{Multimodal Tasks}
We evaluate our model on four multimodal understanding tasks as shown in Table \ref{tab:multimodal-data-info}.
The MMMU-Overall dataset, with 10,500 instances, assesses general multimodal capabilities, while the MMMU-Business dataset, with 1,428 instances, evaluates performance in financial domains such as accounting and marketing.
In addition to existing datasets, in this paper, we build two new financial multi-modal evaluation tasks.
First, \textbf{ChartBench} tests chart interpretation skills with 350 financially relevant instances selected using the \href{https://huggingface.co/datasets/lewy666/ChartInstructionData}{ChartInstructionData}. We employed GPT-4o to assign a finance-relevance score, selecting instances with scores of 9 or above (out of 10). These were categorized into seven groups, as detailed in Appendix~\ref{appendix-tasks}, with 50 randomly chosen instances per category forming the ChartBench benchmark.
The \textbf{TableBench} assesses multimodal capabilities using tabular images, offering a realistic testbed for handling complex financial data. Our dataset includes 450 questions split between comparison and data retrieval tasks, essential for extracting data points and comparing metrics, reflecting key decision-making processes in finance.

\textbf{Performance} 
As shown in Table~\ref{tab:multimodal_results}, our multimodal model, FinLLaVA, our multimodal model, FinLLaVA, achieves the best performance across all tasks among open-source models with 7B and 13B model sizes. FinLLaVA even outperforms larger models like LLaVA-1.5 and LLaVA-1.6, both of which use Vicuna-13B.
On TableBench, FinLLaVA achieves the best performance and outperforms SOTA commercialized LLMs GPT-4 and Gemini-1.5-pro, which proves the effectiveness of our multimodal extension.
These results highlight the robustness and promising performance of FinLLaVA
On the TableBench dataset, FinLLaVA not only achieves the best performance but also surpasses state-of-the-art commercialized LLMs like GPT-4 and Gemini-1.5-pro. This success demonstrates the effectiveness of our multimodal extension in enhancing the model's capability to process and analyze complex financial data. These results highlight its potential for widespread application in the financial domain, offering an efficient solution for interpreting and managing multimodal financial data.
Our evaluation differs from FinTral's in using out-of-domain data, showcasing our model's robustness and superior performance. We enhance its ability to interpret financial tables by integrating extensive OCR data during alignment. With entirely image-based input, users only need to provide table or chart images, making it convenient for real-world applications like financial reporting and auditing. This simplicity allows financial professionals to efficiently use our model, streamlining their workflow and improving productivity without requiring extensive technical knowledge.

\section{Conclusion}
In this paper, we present Open-FinLLMs, an innovative suite of open-source financial language models specifically designed to address the limitations of LLMs in financial applications. Our contributions include FinLLaMA, a foundational model built on the continual pre-training of LLaMA3 8B with an extensive multimodal domain-specific datasets, FinLLaMA-Instruct, fine-tuned with diverse instructions for improved instruction-following and conversational capabilities, and FinLLaVA, the multimodal extension capable of handling vision, tabular, and charts.
Our comprehensive evaluation, which covers 14 financial tasks with $30$ datasets and 4 multimodal tasks, demonstrates that Open-FinLLMs models outperform existing financial LLMs and even better performance to GPT-4 and GPT-4o despite their smaller size, on financial NLP, reasoning, decision-making and multimodal tasks in zero-shot, few-shot, and supervised fine-tuning settings.
The results underscore the potential of Open-FinLLMs in advancing Financial AI applications through their robust performance in various financial tasks.

\section*{Limitation}
Our Open-FinLLMs, including the foundational FinLLaMA, the instruction-tuned FinLLaMA-Instruct, and the multimodal FinLLaVA, demonstrate promising capabilities but have several limitations. First, the models are currently limited to a size of 8B parameters, and future work should explore both smaller models for efficiency and larger models for enhanced performance. Second, our experiments focused exclusively on English, highlighting the need to expand to multilingual settings to better serve global financial markets. Third, the tasks we addressed, such as trading-related scenarios, represent only a subset of potential applications, and future research should investigate broader industrial use cases like financial auditing, risk management, and regulatory compliance. Finally, the multimodal capabilities of FinLLaVA are restricted to handling charts and tabular data; incorporating other data types.

\section*{Ethical Statement}


The development and dissemination of the Open FinLLMs by the authors carry full responsibility for any potential violation of rights or arising legal issues. All raw data we used are publicly available and do not contain any personal information. Diligent efforts have been undertaken to ensure the construction of the Open FinLLMs respects privacy and conforms to established ethical guidelines. The datasets compiled within Open FinLLMs are shared under the MIT license, with the expectation that users agree to adhere to its conditions.

This manuscript, including any associated large language models, source codes, datasets, and appendices ("Material"), is intended solely for academic research and educational purposes. Users must recognize that the Material is not designed to provide financial, legal, medical, or other professional advice, nor should it serve as a substitute for expert judgment or decision-making in any domain.

While the authors have taken reasonable measures to ensure the accuracy and reliability of the Material, no guarantees—explicit or implied—are provided regarding its completeness, accuracy, or fitness for specific purposes. The authors and their affiliated institutions disclaim responsibility for any outcomes, whether direct or indirect, arising from the use, misuse, or reliance upon the Material. Users are strongly advised to consult relevant professionals when making decisions based on outputs generated by this Material.

By accessing or utilizing this Material, individuals agree to indemnify, defend, and hold harmless the authors, as well as their affiliated organizations, from any claims, liabilities, or damages resulting from the use or interpretation of the Material, including outputs produced by the large language model. Users bear the responsibility for ensuring the appropriate and ethical use of this Material in accordance with applicable laws and regulations.

\bibliography{custom}

\begin{thebibliography}{61}
\providecommand{\natexlab}[1]{#1}

\bibitem[{Achiam et~al.(2023)Achiam, Adler, Agarwal, Ahmad, Akkaya, Aleman, Almeida, Altenschmidt, Altman, Anadkat et~al.}]{achiam2023gpt}
Josh Achiam, Steven Adler, Sandhini Agarwal, Lama Ahmad, Ilge Akkaya, Florencia~Leoni Aleman, Diogo Almeida, Janko Altenschmidt, Sam Altman, Shyamal Anadkat, and 1 others. 2023.
\newblock Gpt-4 technical report.
\newblock \emph{arXiv preprint arXiv:2303.08774}.

\bibitem[{Alvarado et~al.(2015)Alvarado, Verspoor, and Baldwin}]{alvarado2015domain}
Julio Cesar~Salinas Alvarado, Karin Verspoor, and Timothy Baldwin. 2015.
\newblock Domain adaption of named entity recognition to support credit risk assessment.
\newblock In \emph{Proceedings of the Australasian Language Technology Association Workshop 2015}, pages 84--90.

\bibitem[{Bhatia et~al.(2024)Bhatia, Nagoudi, Cavusoglu, and Abdul-Mageed}]{bhatia2024fintral}
Gagan Bhatia, El~Moatez~Billah Nagoudi, Hasan Cavusoglu, and Muhammad Abdul-Mageed. 2024.
\newblock Fintral: A family of gpt-4 level multimodal financial large language models.
\newblock \emph{arXiv preprint arXiv:2402.10986}.

\bibitem[{Brown et~al.(2020)Brown, Mann, Ryder, Subbiah, Kaplan, Dhariwal, Neelakantan, Shyam, Sastry, Askell et~al.}]{brown2020language}
Tom Brown, Benjamin Mann, Nick Ryder, Melanie Subbiah, Jared~D Kaplan, Prafulla Dhariwal, Arvind Neelakantan, Pranav Shyam, Girish Sastry, Amanda Askell, and 1 others. 2020.
\newblock Language models are few-shot learners.
\newblock \emph{Advances in Neural Information Processing Systems}, 33:1877--1901.

\bibitem[{Bubeck et~al.(2023)Bubeck, Chandrasekaran, Eldan, Gehrke, Horvitz, Kamar, Lee, Lee, Li, Lundberg et~al.}]{bubeck2023sparks}
S{\'e}bastien Bubeck, Varun Chandrasekaran, Ronen Eldan, Johannes Gehrke, Eric Horvitz, Ece Kamar, Peter Lee, Yin~Tat Lee, Yuanzhi Li, Scott Lundberg, and 1 others. 2023.
\newblock Sparks of artificial general intelligence: Early experiments with {GPT-4}.
\newblock \emph{arXiv preprint arXiv:2303.12712}.

\bibitem[{Chen et~al.(2023)Chen, Lin, Chiu, Huang, Alhamzeh, Huang, Takamura, and Chen}]{chen2023overview}
Chung-Chi Chen, Chin-Yi Lin, Chr-Jr Chiu, Hen-Hsen Huang, Alaa Alhamzeh, Yu-Lieh Huang, Hiroya Takamura, and Hsin-Hsi Chen. 2023.
\newblock Overview of the ntcir-17 finarg-1 task: Fine-grained argument understanding in financial analysis.
\newblock In \emph{Proceedings of the 17th NTCIR Conference on Evaluation of Information Access Technologies, Tokyo, Japan}, pages 12--15.

\bibitem[{Chen et~al.(2024{\natexlab{a}})Chen, Huang, Ma, Chen, Pan, Ge, Gao, Xie, Liu, Gao, Li, Ding, and Zhou}]{chen2024datajuicer}
Daoyuan Chen, Yilun Huang, Zhijian Ma, Hesen Chen, Xuchen Pan, Ce~Ge, Dawei Gao, Yuexiang Xie, Zhaoyang Liu, Jinyang Gao, Yaliang Li, Bolin Ding, and Jingren Zhou. 2024{\natexlab{a}}.
\newblock Data-juicer: A one-stop data processing system for large language models.
\newblock In \emph{International Conference on Management of Data}.

\bibitem[{Chen et~al.(2024{\natexlab{b}})Chen, Chen, Zhang, Chen, Wu, Zhang, Chen, Li, Wan, and Wang}]{chen2024allava}
Guiming~Hardy Chen, Shunian Chen, Ruifei Zhang, Junying Chen, Xiangbo Wu, Zhiyi Zhang, Zhihong Chen, Jianquan Li, Xiang Wan, and Benyou Wang. 2024{\natexlab{b}}.
\newblock \href {https://arxiv.org/abs/2402.11684} {{ALLaVA}: Harnessing {GPT4V}-synthesized data for a lite vision-language model}.
\newblock \emph{Preprint}, arXiv:2402.11684.

\bibitem[{Chen et~al.(2021)Chen, Chen, Smiley, Shah, Borova, Langdon, Moussa, Beane, Huang, Routledge, and Wang}]{chen2021finqa}
Zhiyu Chen, Wenhu Chen, Charese Smiley, Sameena Shah, Iana Borova, Dylan Langdon, Reema Moussa, Matt Beane, Ting-Hao Huang, Bryan Routledge, and William~Yang Wang. 2021.
\newblock Finqa: A dataset of numerical reasoning over financial data.
\newblock \emph{Proceedings of EMNLP 2021}.

\bibitem[{Chen et~al.(2022)Chen, Li, Smiley, Ma, Shah, and Wang}]{chen2022convfinqa}
Zhiyu Chen, Shiyang Li, Charese Smiley, Zhiqiang Ma, Sameena Shah, and William~Yang Wang. 2022.
\newblock \href {https://arxiv.org/abs/2210.03849} {Convfinqa: Exploring the chain of numerical reasoning in conversational finance question answering}.
\newblock \emph{Preprint}, arXiv:2210.03849.

\bibitem[{Cortis et~al.(2017)Cortis, Freitas, Daudert, Huerlimann, Zarrouk, Handschuh, and Davis}]{cortis2017semeval}
Keith Cortis, Andr{\'e} Freitas, Tobias Daudert, Manuela Huerlimann, Manel Zarrouk, Siegfried Handschuh, and Brian Davis. 2017.
\newblock Semeval-2017 task 5: Fine-grained sentiment analysis on financial microblogs and news.
\newblock In \emph{Proceedings of the 11th international workshop on semantic evaluation (SemEval-2017)}, pages 519--535.

\bibitem[{Dettmers et~al.(2024)Dettmers, Pagnoni, Holtzman, and Zettlemoyer}]{dettmers2024qlora}
Tim Dettmers, Artidoro Pagnoni, Ari Holtzman, and Luke Zettlemoyer. 2024.
\newblock Qlora: Efficient finetuning of quantized llms.
\newblock \emph{Advances in Neural Information Processing Systems}, 36.

\bibitem[{Feng et~al.(2024)Feng, Dai, Huang, Zhang, Xie, Han, Chen, Lopez-Lira, and Wang}]{feng2024empowering}
Duanyu Feng, Yongfu Dai, Jimin Huang, Yifang Zhang, Qianqian Xie, Weiguang Han, Zhengyu Chen, Alejandro Lopez-Lira, and Hao Wang. 2024.
\newblock \href {https://arxiv.org/abs/2310.00566} {Empowering many, biasing a few: Generalist credit scoring through large language models}.
\newblock \emph{Preprint}, arXiv:2310.00566.

\bibitem[{Gupta et~al.()Gupta, Th{\'e}rien, Ibrahim, Richter, Anthony, Belilovsky, Rish, and Lesort}]{guptacontinual}
Kshitij Gupta, Benjamin Th{\'e}rien, Adam Ibrahim, Mats~Leon Richter, Quentin~Gregory Anthony, Eugene Belilovsky, Irina Rish, and Timoth{\'e}e Lesort.
\newblock Continual pre-training of large language models: How to re-warm your model?
\newblock In \emph{Workshop on Efficient Systems for Foundation Models@ ICML2023}.

\bibitem[{Hofmann(1994)}]{misc_german_credit_data_144}
Hans Hofmann. 1994.
\newblock {Statlog (German Credit Data)}.
\newblock UCI Machine Learning Repository.
\newblock {DOI}: https://doi.org/10.24432/C5NC77.

\bibitem[{Jiang et~al.(2023)Jiang, Sablayrolles, Mensch, Bamford, Chaplot, Casas, Bressand, Lengyel, Lample, Saulnier et~al.}]{jiang2023mistral}
Albert~Q Jiang, Alexandre Sablayrolles, Arthur Mensch, Chris Bamford, Devendra~Singh Chaplot, Diego de~las Casas, Florian Bressand, Gianna Lengyel, Guillaume Lample, Lucile Saulnier, and 1 others. 2023.
\newblock Mistral 7b.
\newblock \emph{arXiv preprint arXiv:2310.06825}.

\bibitem[{Liu et~al.(2024{\natexlab{a}})Liu, Li, Li, and Lee}]{liu2023improved}
Haotian Liu, Chunyuan Li, Yuheng Li, and Yong~Jae Lee. 2024{\natexlab{a}}.
\newblock Improved baselines with visual instruction tuning.
\newblock In \emph{IEEE/CVF Conference on Computer Vision and Pattern Recognition}, pages 26296--26306.

\bibitem[{Liu et~al.(2024{\natexlab{b}})Liu, Li, Wu, and Lee}]{liu2023llava}
Haotian Liu, Chunyuan Li, Qingyang Wu, and Yong~Jae Lee. 2024{\natexlab{b}}.
\newblock Visual instruction tuning.
\newblock \emph{Advances in Neural Information Processing Systems}, 36.

\bibitem[{Liu et~al.(2023)Liu, Wang, Yang, and Zha}]{liu2023fingpt}
Xiao-Yang Liu, Guoxuan Wang, Hongyang Yang, and Daochen Zha. 2023.
\newblock {Data-centric FinGPT}: Democratizing internet-scale data for financial large language models.
\newblock \emph{Workshop on Instruction Tuning and Instruction Following, NeurIPS}.

\bibitem[{Liu et~al.(2024{\natexlab{c}})Liu, Zhang, Wang, Tong, and Walid}]{liu2024fingpt}
Xiao-Yang Liu, Jie Zhang, Guoxuan Wang, Weiqing Tong, and Anwar Walid. 2024{\natexlab{c}}.
\newblock {FinGPT-HPC}: Efficient pretraining and finetuning large language models for financial applications with high-performance computing.
\newblock \emph{arXiv preprint arXiv:2402.13533}.

\bibitem[{Loshchilov and Hutter()}]{loshchilovdecoupled}
Ilya Loshchilov and Frank Hutter.
\newblock Decoupled weight decay regularization.
\newblock In \emph{International Conference on Learning Representations}.

\bibitem[{Maia et~al.(2018)Maia, Handschuh, Freitas, Davis, McDermott, Zarrouk, and Balahur}]{maia2018www}
Macedo Maia, Siegfried Handschuh, Andre Freitas, Brian Davis, Ross McDermott, Manel Zarrouk, and Alexandra Balahur. 2018.
\newblock \href {https://doi.org/10.1145/3184558.3192301} {Www'18 open challenge: Financial opinion mining and question answering}.

\bibitem[{Malo et~al.(2014)Malo, Sinha, Korhonen, Wallenius, and Takala}]{malo2014good}
Pekka Malo, Ankur Sinha, Pekka Korhonen, Jyrki Wallenius, and Pyry Takala. 2014.
\newblock Good debt or bad debt: Detecting semantic orientations in economic texts.
\newblock \emph{Journal of the Association for Information Science and Technology}, 65(4):782--796.

\bibitem[{Mariko et~al.(2020)Mariko, Akl, Labidurie, Durfort, De~Mazancourt, and El-Haj}]{mariko2020financial}
Dominique Mariko, Hanna~Abi Akl, Estelle Labidurie, Stephane Durfort, Hugues De~Mazancourt, and Mahmoud El-Haj. 2020.
\newblock Financial document causality detection shared task (fincausal 2020).
\newblock \emph{arXiv preprint arXiv:2012.02505}.

\bibitem[{Masry et~al.(2023)Masry, Kavehzadeh, Hoque, Joty et~al.}]{masry2023unichart}
Ahmed Masry, Parsa Kavehzadeh, Enamul Hoque, Shafiq Joty, and 1 others. 2023.
\newblock Unichart: A universal vision-language pretrained model for chart comprehension and reasoning.
\newblock In \emph{The 2023 Conference on Empirical Methods in Natural Language Processing}.

\bibitem[{Masry et~al.(2022)Masry, Long, Tan, Joty, and Hoque}]{masry-etal-2022-chartqa}
Ahmed Masry, Do~Long, Jia~Qing Tan, Shafiq Joty, and Enamul Hoque. 2022.
\newblock {C}hart{QA}: A benchmark for question answering about charts with visual and logical reasoning.
\newblock In \emph{Findings of the Association for Computational Linguistics: ACL 2022}, pages 2263--2279, Dublin, Ireland. Association for Computational Linguistics.

\bibitem[{Mukherjee et~al.(2022)Mukherjee, Bohra, Banerjee, Sharma, Hegde, Shaikh, Shrivastava, Dasgupta, Ganguly, Ghosh et~al.}]{mukherjee2022ectsum}
Rajdeep Mukherjee, Abhinav Bohra, Akash Banerjee, Soumya Sharma, Manjunath Hegde, Afreen Shaikh, Shivani Shrivastava, Koustuv Dasgupta, Niloy Ganguly, Saptarshi Ghosh, and 1 others. 2022.
\newblock Ectsum: A new benchmark dataset for bullet point summarization of long earnings call transcripts.
\newblock \emph{arXiv preprint arXiv:2210.12467}.

\bibitem[{Nassar et~al.(2022)Nassar, Livathinos, Lysak, and Staar}]{nassar2022tableformer}
Ahmed Nassar, Nikolaos Livathinos, Maksym Lysak, and Peter Staar. 2022.
\newblock Tableformer: Table structure understanding with transformers.
\newblock In \emph{IEEE/CVF Conference on Computer Vision and Pattern Recognition}, pages 4614--4623.

\bibitem[{Nie et~al.(2024)Nie, Kong, Dong, Mulvey, Poor, Wen, and Zohren}]{nie2024survey}
Yuqi Nie, Yaxuan Kong, Xiaowen Dong, John~M Mulvey, H~Vincent Poor, Qingsong Wen, and Stefan Zohren. 2024.
\newblock A survey of large language models for financial applications: Progress, prospects and challenges.
\newblock \emph{arXiv preprint arXiv:2406.11903}.

\bibitem[{Obeid and Hoque(2020)}]{obeid2020charttotextgeneratingnaturallanguage}
Jason Obeid and Enamul Hoque. 2020.
\newblock Chart-to-text: Generating natural language descriptions for charts by adapting the transformer model.
\newblock In \emph{International Conference on Natural Language Generation}, pages 138--147.

\bibitem[{OpenAI(2023)}]{openai2023gpt4}
OpenAI. 2023.
\newblock \href {https://arxiv.org/abs/2303.08774} {Gpt-4 technical report}.
\newblock \emph{Preprint}, arXiv:2303.08774.

\bibitem[{Penedo et~al.(2024)Penedo, Kydl{\'\i}{\v{c}}ek, Lozhkov, Mitchell, Raffel, Von~Werra, Wolf et~al.}]{penedo2024fineweb}
Guilherme Penedo, Hynek Kydl{\'\i}{\v{c}}ek, Anton Lozhkov, Margaret Mitchell, Colin Raffel, Leandro Von~Werra, Thomas Wolf, and 1 others. 2024.
\newblock The {FineWeb} datasets: Decanting the web for the finest text data at scale.
\newblock \emph{arXiv preprint arXiv:2406.17557}.

\bibitem[{Qian et~al.(2025)Qian, Zhou, Wang, Peng, Huang, and Xie}]{qian2025fino1}
Lingfei Qian, Weipeng Zhou, Yan Wang, Xueqing Peng, Jimin Huang, and Qianqian Xie. 2025.
\newblock Fino1: On the transferability of reasoning enhanced llms to finance.
\newblock \emph{arXiv preprint arXiv:2502.08127}.

\bibitem[{Quinlan()}]{misc_australian_credit_approval_143}
Ross Quinlan.
\newblock {Statlog (Australian Credit Approval)}.
\newblock UCI Machine Learning Repository.
\newblock {DOI}: https://doi.org/10.24432/C59012.

\bibitem[{Radford et~al.(2021)Radford, Kim, Hallacy, Ramesh, Goh, Agarwal, Sastry, Askell, Mishkin, Clark, Krueger, and Sutskever}]{radford2021learningtransferablevisualmodels}
Alec Radford, Jong~Wook Kim, Chris Hallacy, Aditya Ramesh, Gabriel Goh, Sandhini Agarwal, Girish Sastry, Amanda Askell, Pamela Mishkin, Jack Clark, Gretchen Krueger, and Ilya Sutskever. 2021.
\newblock \href {https://arxiv.org/abs/2103.00020} {Learning transferable visual models from natural language supervision}.
\newblock \emph{Preprint}, arXiv:2103.00020.

\bibitem[{Rangapur et~al.(2023)Rangapur, Wang, and Shu}]{rangapur2023fin}
Aman Rangapur, Haoran Wang, and Kai Shu. 2023.
\newblock Fin-fact: A benchmark dataset for multimodal financial fact checking and explanation generation.
\newblock \emph{arXiv preprint arXiv:2309.08793}.

\bibitem[{Roziere et~al.(2023)Roziere, Gehring, Gloeckle, Sootla, Gat, Tan, Adi, Liu, Remez, Rapin et~al.}]{roziere2023code}
Baptiste Roziere, Jonas Gehring, Fabian Gloeckle, Sten Sootla, Itai Gat, Xiaoqing~Ellen Tan, Yossi Adi, Jingyu Liu, Tal Remez, J{\'e}r{\'e}my Rapin, and 1 others. 2023.
\newblock Code llama: Open foundation models for code.
\newblock \emph{arXiv preprint arXiv:2308.12950}.

\bibitem[{Shah et~al.(2023{\natexlab{a}})Shah, Paturi, and Chava}]{shah2023trillion}
Agam Shah, Suvan Paturi, and Sudheer Chava. 2023{\natexlab{a}}.
\newblock \href {https://doi.org/10.18653/v1/2023.acl-long.368} {Trillion dollar words: A new financial dataset, task {\&} market analysis}.
\newblock In \emph{Proceedings of the 61st Annual Meeting of the Association for Computational Linguistics (Volume 1: Long Papers)}, pages 6664--6679, Toronto, Canada. Association for Computational Linguistics.

\bibitem[{Shah et~al.(2023{\natexlab{b}})Shah, Vithani, Gullapalli, and Chava}]{shah2023finer}
Agam Shah, Ruchit Vithani, Abhinav Gullapalli, and Sudheer Chava. 2023{\natexlab{b}}.
\newblock Finer: Financial named entity recognition dataset and weak-supervision model.
\newblock \emph{arXiv preprint arXiv:2302.11157}.

\bibitem[{Sinha and Khandait(2021)}]{sinha2021impact}
Ankur Sinha and Tanmay Khandait. 2021.
\newblock Impact of news on the commodity market: Dataset and results.
\newblock In \emph{Advances in Information and Communication: Proceedings of the 2021 Future of Information and Communication Conference (FICC), Volume 2}, pages 589--601. Springer.

\bibitem[{Soun et~al.(2022)Soun, Yoo, Cho, Jeon, and Kang}]{soun2022accurate}
Yejun Soun, Jaemin Yoo, Minyong Cho, Jihyeong Jeon, and U~Kang. 2022.
\newblock Accurate stock movement prediction with self-supervised learning from sparse noisy tweets.
\newblock In \emph{2022 IEEE International Conference on Big Data (Big Data)}, pages 1691--1700. IEEE.

\bibitem[{Touvron et~al.(2023{\natexlab{a}})Touvron, Lavril, Izacard, Martinet, Lachaux, Lacroix, Rozière, Goyal, Hambro, Azhar, Rodriguez, Joulin, Grave, and Lample}]{touvron2023llama}
Hugo Touvron, Thibaut Lavril, Gautier Izacard, Xavier Martinet, Marie-Anne Lachaux, Timothée Lacroix, Baptiste Rozière, Naman Goyal, Eric Hambro, Faisal Azhar, Aurelien Rodriguez, Armand Joulin, Edouard Grave, and Guillaume Lample. 2023{\natexlab{a}}.
\newblock \href {https://arxiv.org/abs/2302.13971} {Llama: Open and efficient foundation language models}.
\newblock \emph{Preprint}, arXiv:2302.13971.

\bibitem[{Touvron et~al.(2023{\natexlab{b}})Touvron, Martin, Stone, Albert, Almahairi, Babaei, Bashlykov et~al.}]{touvron2023llama2openfoundation}
Hugo Touvron, Louis Martin, Kevin Stone, Peter Albert, Amjad Almahairi, Yasmine Babaei, Nikolay Bashlykov, and 1 others. 2023{\natexlab{b}}.
\newblock \href {https://arxiv.org/abs/2307.09288} {Llama 2: Open foundation and fine-tuned chat models}.
\newblock \emph{Preprint}, arXiv:2307.09288.

\bibitem[{Wang et~al.(2023)Wang, Meng, Weng, He, Wu, and Jiang}]{wang2023believepromptinggpt4vbetter}
Junke Wang, Lingchen Meng, Zejia Weng, Bo~He, Zuxuan Wu, and Yu-Gang Jiang. 2023.
\newblock \href {https://arxiv.org/abs/2311.07574} {To see is to believe: Prompting gpt-4v for better visual instruction tuning}.
\newblock \emph{Preprint}, arXiv:2311.07574.

\bibitem[{Wu et~al.(2018)Wu, Zhang, Shen, and Wang}]{wu2018hybrid}
Huizhe Wu, Wei Zhang, Weiwei Shen, and Jun Wang. 2018.
\newblock Hybrid deep sequential modeling for social text-driven stock prediction.
\newblock In \emph{Proceedings of the 27th ACM international conference on information and knowledge management}, pages 1627--1630.

\bibitem[{Wu et~al.(2023{\natexlab{a}})Wu, Irsoy, Lu, Dabravolski, Dredze, Gehrmann, Kambadur, Rosenberg, and Mann}]{wu2023bloomberggpt}
Shijie Wu, Ozan Irsoy, Steven Lu, Vadim Dabravolski, Mark Dredze, Sebastian Gehrmann, Prabhanjan Kambadur, David Rosenberg, and Gideon Mann. 2023{\natexlab{a}}.
\newblock \href {https://arxiv.org/abs/2303.17564} {Bloomberggpt: A large language model for finance}.
\newblock \emph{Preprint}, arXiv:2303.17564.

\bibitem[{Wu et~al.(2023{\natexlab{b}})Wu, Irsoy, Lu, Dabravolski, Dredze, Gehrmann, Kambadur, Rosenberg, and Mann}]{wu2023bloomberggptlargelanguagemodel}
Shijie Wu, Ozan Irsoy, Steven Lu, Vadim Dabravolski, Mark Dredze, Sebastian Gehrmann, Prabhanjan Kambadur, David Rosenberg, and Gideon Mann. 2023{\natexlab{b}}.
\newblock \href {https://arxiv.org/abs/2303.17564} {{BloombergGPT}: A large language model for finance}.
\newblock \emph{Preprint}, arXiv:2303.17564.

\bibitem[{Wu et~al.(2024)Wu, Luo, Li, Pan, Vu, and Haffari}]{wu2024continual}
Tongtong Wu, Linhao Luo, Yuan-Fang Li, Shirui Pan, Thuy-Trang Vu, and Gholamreza Haffari. 2024.
\newblock Continual learning for large language models: A survey.
\newblock \emph{arXiv preprint arXiv:2402.01364}.

\bibitem[{Xie et~al.(2024{\natexlab{a}})Xie, Han, Chen, Xiang, Zhang, He, Xiao, Li, Dai, Feng et~al.}]{xie2024finben}
Qianqian Xie, Weiguang Han, Zhengyu Chen, Ruoyu Xiang, Xiao Zhang, Yueru He, Mengxi Xiao, Dong Li, Yongfu Dai, Duanyu Feng, and 1 others. 2024{\natexlab{a}}.
\newblock The finben: An holistic financial benchmark for large language models.
\newblock \emph{arXiv preprint arXiv:2402.12659}.

\bibitem[{Xie et~al.(2023{\natexlab{a}})Xie, Han, Lai, Peng, and Huang}]{xie2023wallstreetneophytezeroshot}
Qianqian Xie, Weiguang Han, Yanzhao Lai, Min Peng, and Jimin Huang. 2023{\natexlab{a}}.
\newblock \href {https://arxiv.org/abs/2304.05351} {The wall street neophyte: A zero-shot analysis of chatgpt over multimodal stock movement prediction challenges}.
\newblock \emph{Preprint}, arXiv:2304.05351.

\bibitem[{Xie et~al.(2023{\natexlab{b}})Xie, Han, Zhang, Lai, Peng, Lopez-Lira, and Huang}]{xie2023pixiu}
Qianqian Xie, Weiguang Han, Xiao Zhang, Yanzhao Lai, Min Peng, Alejandro Lopez-Lira, and Jimin Huang. 2023{\natexlab{b}}.
\newblock Pixiu: A large language model, instruction data and evaluation benchmark for finance.
\newblock In \emph{Proceedings of the 37th International Conference on Neural Information Processing Systems}, pages 33469--33484.

\bibitem[{Xie et~al.(2024{\natexlab{b}})Xie, Pham, Dong, Du, Liu, Lu, Liang, Le, Ma, and Yu}]{xie2024doremi}
Sang~Michael Xie, Hieu Pham, Xuanyi Dong, Nan Du, Hanxiao Liu, Yifeng Lu, Percy~S Liang, Quoc~V Le, Tengyu Ma, and Adams~Wei Yu. 2024{\natexlab{b}}.
\newblock Doremi: Optimizing data mixtures speeds up language model pretraining.
\newblock \emph{Advances in Neural Information Processing Systems}, 36.

\bibitem[{Xu and Cohen(2018)}]{xu2018stock}
Yumo Xu and Shay~B Cohen. 2018.
\newblock Stock movement prediction from tweets and historical prices.
\newblock In \emph{Proceedings of the 56th Annual Meeting of the Association for Computational Linguistics (Volume 1: Long Papers)}, pages 1970--1979.

\bibitem[{Yang et~al.(2023)Yang, Liu, and Wang}]{yang2023fingpt}
Hongyang Yang, Xiao-Yang Liu, and Christina~Dan Wang. 2023.
\newblock {FinGPT}: Open-source financial large language models.
\newblock \emph{FinLLM at IJCAI}.

\bibitem[{Yang et~al.(2020)Yang, Kenny, Ng, Yang, Smyth, and Dong}]{yang2020generating}
Linyi Yang, Eoin~M Kenny, Tin Lok~James Ng, Yi~Yang, Barry Smyth, and Ruihai Dong. 2020.
\newblock Generating plausible counterfactual explanations for deep transformers in financial text classification.
\newblock \emph{arXiv preprint arXiv:2010.12512}.

\bibitem[{Yu et~al.(2024)Yu, Li, Chen, Jiang, Li, Zhang, Liu, Suchow, and Khashanah}]{yu2024finmem}
Yangyang Yu, Haohang Li, Zhi Chen, Yuechen Jiang, Yang Li, Denghui Zhang, Rong Liu, Jordan~W Suchow, and Khaldoun Khashanah. 2024.
\newblock {FinMem}: A performance-enhanced llm trading agent with layered memory and character design.
\newblock In \emph{AAAI Symposium Series}, volume~3, pages 595--597.

\bibitem[{Yue et~al.(2024)Yue, Ni, Zhang, Zheng, Liu, Zhang, Stevens, Jiang, Ren, Sun, Wei, Yu, Yuan, Sun, Yin, Zheng, Yang, Liu, Huang, Sun, Su, and Chen}]{yue2023mmmu}
Xiang Yue, Yuansheng Ni, Kai Zhang, Tianyu Zheng, Ruoqi Liu, Ge~Zhang, Samuel Stevens, Dongfu Jiang, Weiming Ren, Yuxuan Sun, Cong Wei, Botao Yu, Ruibin Yuan, Renliang Sun, Ming Yin, Boyuan Zheng, Zhenzhu Yang, Yibo Liu, Wenhao Huang, and 3 others. 2024.
\newblock {MMMU}: A massive multi-discipline multimodal understanding and reasoning benchmark for expert {AGI}.
\newblock In \emph{IEEE/CVF Conference on Computer Vision and Pattern Recognition}.

\bibitem[{Yue et~al.(2023)Yue, Qu, Zhang, Fu, Huang, Sun, Su, and Chen}]{yue2023mammoth}
Xiang Yue, Xingwei Qu, Ge~Zhang, Yao Fu, Wenhao Huang, Huan Sun, Yu~Su, and Wenhu Chen. 2023.
\newblock Mammoth: Building math generalist models through hybrid instruction tuning.
\newblock \emph{arXiv preprint arXiv:2309.05653}.

\bibitem[{Zhao et~al.(2023{\natexlab{a}})Zhao, Liu, Long, Zhang, Zhao, and Cohan}]{zhao2023knowledgemath}
Yilun Zhao, Hongjun Liu, Yitao Long, Rui Zhang, Chen Zhao, and Arman Cohan. 2023{\natexlab{a}}.
\newblock Knowledgemath: Knowledge-intensive math word problem solving in finance domains.
\newblock \emph{arXiv preprint arXiv:2311.09797}.

\bibitem[{Zhao et~al.(2023{\natexlab{b}})Zhao, Long, Liu, Nan, Chen, Kamoi, Liu, Tang, Zhang, and Cohan}]{zhao2023docmath}
Yilun Zhao, Yitao Long, Hongjun Liu, Linyong Nan, Lyuhao Chen, Ryo Kamoi, Yixin Liu, Xiangru Tang, Rui Zhang, and Arman Cohan. 2023{\natexlab{b}}.
\newblock Docmath-eval: Evaluating numerical reasoning capabilities of llms in understanding long documents with tabular data.
\newblock \emph{arXiv preprint arXiv:2311.09805}.

\bibitem[{Zhou et~al.(2021)Zhou, Ma, and Liu}]{zhou2021trade}
Zhihan Zhou, Liqian Ma, and Han Liu. 2021.
\newblock Trade the event: Corporate events detection for news-based event-driven trading.
\newblock \emph{arXiv preprint arXiv:2105.12825}.

\end{thebibliography}

\clearpage
\appendix

\section{Curation of Continual Pre-training Corpus}
\label{data-building}
\subsection{Dataset Details}
Our continual pre-training corpus is designed to ensure comprehensive coverage of financial knowledge by integrating a diverse range of data sources. This appendix provides a detailed overview of each data source:
\begin{itemize}
    \item {Financial Papers}: This subset includes 4 billion tokens extracted from academic papers and research articles, offering a strong foundation in financial concepts and theories. The papers span a period from 2000 to 2023. These documents are sourced from SSRN and open-source conference proceedings, providing in-depth insights into both foundational and cutting-edge financial research.
    \item Conference Calls: Comprising 5 billion tokens, this dataset includes open-source transcripts from earnings calls, analyst meetings, and investor briefings, collected from 09/08/2004 to 12/17/2021. These transcripts provide real-time insights into corporate performance and strategic directions, allowing for a nuanced understanding of company operations and market positioning. Sources include major corporations across various industries, reflecting a diverse set of perspectives and strategies.
    \item Financial Reports: This component consists of 5 billion tokens from annual and quarterly reports, covering the period from 2005 to 2020. These reports are crucial for assessing a company's financial health, market positioning, and strategic outlook. They include balance sheets, income statements, and management discussions, providing a comprehensive view of corporate financial performance.
    \item Technical Indicators: With 12 billion tokens, this dataset includes open-source economic indicators and financial ratios sourced from company filings and open market data, spanning from 2009 to 2023. These indicators are essential for macroeconomic analysis and investment decision-making, covering metrics such as GDP, inflation rates, interest rates, and key financial ratios.
    \item News and Social Media: This subset includes 7 billion tokens from financial news outlets and social media platforms, collected from 1999 to 2021. This data provides timely updates on market trends, public opinion, and emerging issues, reflecting the dynamic nature of financial markets. Sources include leading financial news websites, and financial forums, capturing both traditional media and real-time public sentiment.
    \item Historical Data: Encompassing 13 billion tokens, this dataset includes historical stock prices, trading volumes, and market data from 1999 to 2022. This data is vital for quantitative analysis and algorithmic trading, providing historical context and trend analysis capabilities. The data is sourced from Yahoo Finance, offering a robust foundation for time-series analysis and predictive modeling.
    \item SEC Filings: This section includes 6 billion tokens from U.S. SEC filings, such as 10-K and 10-Q reports, spanning from 1994 to 2020. These filings provide comprehensive insights into corporate activities, financial conditions, and risk factors. They are sourced from the U.S. Securities and Exchange Commission's EDGAR database, ensuring official and up-to-date corporate information.
\end{itemize}
\subsubsection{Data Processing \& Cleaning}
In our data preprocessing pipeline for training Open-FinLLMs, we utilize Data-Juicer~\citep{chen2024datajuicer} to clean and standardize the datasets. For each corpus, we remove email addresses and URLs to enhance privacy and reduce noise, ensuring the focus remains on the textual content. We address unicode inconsistencies by standardizing characters across the dataset, which maintains uniformity and aids in accurate text representation. Punctuation is normalized to provide consistency in text parsing, while excess whitespace is removed to improve readability and structure. For tabular and time-series data, we first split them into rows into samples of approximately 2,048 tokens each, formatting each block in HTML and ensuring each includes the table header for context.
We then combine all datasets and further chunk the entire dataset into 8,192 token blocks, readying the data for efficient processing by the model.

\subsection{Tabular and Time-series Data Format}
\begin{center}
\fcolorbox{black}{gray!10}{\parbox{.95\linewidth}{
\textless{}table\textgreater{}

  \textless{}thead\textgreater{}
    \textless{}tr\textgreater{}\textless{}th\textgreater{}Column Header 1\textless{}/th\textgreater{}\textless{}th\textgreater{}Column Header 2\textless{}/th\textgreater{}\textless{}/tr\textgreater{}

  \textless{}/thead\textgreater{}

  \textless{}tbody\textgreater{}
    \textless{}tr\textgreater{}\textless{}td\textgreater{}Data Row 1, Cell 1\textless{}/td\textgreater{}\textless{}td\textgreater{}Data Row 1, Cell 2\textless{}/td\textgreater{}\textless{}/tr\textgreater{}

  \textless{}/tbody\textgreater{}

\textless{}/table\textgreater{}
}
}
\end{center}

\section{Financial Instruction Dataset}
\label{instruction-data}
\subsection{Data Sources}
Our financial instruction dataset is a compilation of diverse and specialized datasets designed to enhance the capabilities of Open-FinLLMs. Below, we provide detailed descriptions of each dataset and the specific tasks they cover:
\begin{itemize}
    \item \textbf{ChanceFocus/FLUPE~\citep{xie2023pixiu}}: The FLUPE dataset, is instrumental in improving financial natural language processing capabilities. It includes tasks such as financial sentiment analysis, news headline classification, and named entity recognition (NER). These tasks involve analyzing financial texts to identify sentiment, classify financial headlines, and recognize entities within financial documents. By providing diverse examples, FLUPE helps models refine their understanding of financial language and context, supporting improved task performance across various financial NLP applications.
    \textbf{FinGPT/Fingpt-finred~\citep{liu2023fingpt}}: This dataset focuses on financial report and document comprehension, with approximately 32.67k examples. It is designed to enhance the model's ability to interpret complex financial documents. The dataset enables models to improve their analytical skills and decision-making abilities based on comprehensive document analysis.
    \item \textbf{TIGER-Lab/MathInstruct~\citep{yue2023mammoth}}: Comprising 262k examples, the MathInstruct dataset is built from 13 distinct mathematical rationale datasets. It includes tasks such as arithmetic operations, algebraic reasoning, probability calculations, statistical analysis, and calculus-based problem solving. The dataset employs methods like chain-of-thought (CoT) and program-of-thought (PoT) rationales to provide intermediate reasoning capabilities across these mathematical fields. This is crucial for financial tasks that require precise calculations and quantitative insights, enabling models to tackle mathematical problems effectively within financial contexts.
    \item \textbf{sujet-ai/Sujet-Finance-Instruct-177k\footnote{\url{https://huggingface.co/datasets/sujet-ai/Sujet-Finance-Instruct-177k}}}: The Sujet-Finance-Instruct-177k dataset is a comprehensive collection of financial textual data, designed for fine-tuning language learning models for specialized financial tasks. It integrates data from 18 different datasets, providing a total of 177,597 entries. The dataset covers a wide range of financial tasks, including:
    \begin{itemize}
        \item Sentiment Analysis: 44,209 entries focused on categorizing financial texts into sentiments such as positive, negative, neutral, bearish, or bullish.
        \item Question Answering (QA): 38,801 entries for direct-answer financial questions that do not require additional context.
        \item QA with Context: 40,475 entries where financial questions require contextual understanding for accurate answers.
        \item QA Conversation: 15,613 entries involving conversational interactions between a user and an LLM assistant.
        \item Yes/No Questions: 20,547 entries focused on questions necessitating a simple yes or no answer.
        \item Topic Classification: 16,990 entries for classifying financial texts into specific finance-related categories.
        \item NER Sentiment Analysis: 962 entries for conducting sentiment analysis at the entity level within texts.
    \end{itemize}
\end{itemize}

\subsection{Data Processing}
In our instruction dataset, we identified overlapping task samples between the FLUPE and Sujet-Finance-Instruct-177k datasets, particularly in tasks such as sentiment analysis and NER. To address this, we manually excluded these redundant samples, which resulted in the removal of approximately 30,000 samples. This step was crucial to ensure that each task is represented uniquely and effectively in the dataset, avoiding any biases that could arise from duplicated entries.

\section{Multimodal Instruction Data} 
\label{appendix-prompt}

\subsection{Table}
\label{table}

Our table data is sourced from the \textit{Fintabnet} and \textit{Marketing} categories of \textit{SynthTabNet}, featuring real financial and marketing tables with diverse layouts. Each table includes parsed bounding boxes, allowing us to reconstruct structure-aware prompts, which are more accurate than image-based descriptions.

\begin{figure*}
\caption{
    Example table image.\\
    \begin{minipage}[t]{1\textwidth}
        \textcolor{blue}{Description:}\\
        \textit{Q: What financial metrics are represented in the table, and how are they organized?}\\
        \textit{A:} The table presents various financial metrics organized into rows and columns, with headers indicating the type of data. The first column lists categories such as ``Corporate'', ``For the year ended December 31'', and ``General and administrative expenses''. The corresponding numerical values in the adjacent columns include 628,856, 518,246, 222,216, and 793,595, respectively. The data appears to be structured in a way that highlights different financial activities and their associated values, with totals and segment information also included. For instance, the ``Total segment'' row shows a value of 886,238, while ``Cash and cash equivalents at beginning of period'' has a value of 946,929. The table also indicates that the data is presented in millions of dollars, emphasizing the scale of the financial figures.\\
    \end{minipage}
    \hfill
    \begin{minipage}[t]{1\textwidth}
        \textcolor{blue}{Instruction (Make Comparisons):}\\
        \textit{Q: Which segment had the highest ``Total net revenues'' and what was the amount?}\\
        \textit{A:} The segment with the highest ``Total net revenues'' is ``Millions of dollars'' with an amount of 972,582.\\
    \end{minipage}
}
\label{fig:chart-example}
\centering
    \includegraphics[width=1\textwidth]{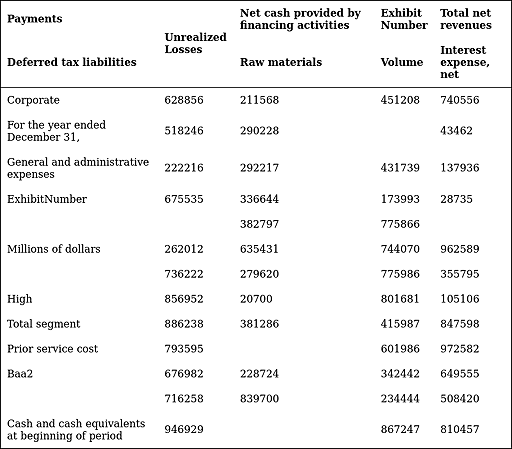}
\end{figure*}

To ensure OCR quality, we limit tables to a maximum size of $10 \times 10$ to avoid resolution-related cell blurring. In the SFT stage, we design seven financial-specific tasks (Appendix~\ref{appendix-tasks}), randomly sampled in the prompts. Figure~\ref{fig:chart-example} shows an example table and how we align and generate SFT data.

\subsection{Chart}
\label{chart}

\begin{figure}
\centering
\begin{minipage}[t]{1\textwidth}
    \centering
    \caption{
        Example chart image~1.\\
        \textcolor{blue}{Description:}\\
        \textit{Q: Describe the type of information that the chart provides.}\\
        \textit{A:} The chart shows year-over-year growth rates for various countries in 2014 and 2015. Each country has two bars representing growth rates for 2014 (in dark blue) and 2015 (in light blue). The growth rates are measured in percentages and are displayed on the horizontal axis, which ranges from -7.5\% to 20\%. The countries listed on the left axis include China, Russia, Germany, India, South Korea, United States, Brazil, France, Australia, Canada, United Kingdom, Spain, Netherlands, Italy, and Japan.\\
        Key information includes:\\
        China's growth rates were 17.8\% in 2014 and 16.5\% in 2015.\\
        Russia had a negative growth rate of -5.4\% in 2014 and a positive growth rate of 13.6\% in 2015.\\
        Germany's growth rates were 7\% in 2014 and 10.6\% in 2015.\\
        India's growth rates were 2.1\% in 2014 and 7.6\% in 2015.\\
        South Korea had growth rates of 5.9\% in 2014 and 5.8\% in 2015.\\
        The United States had growth rates of 5.9\% in 2014 and 6.8\% in 2015.\\
        Brazil had growth rates of 5.9\% in 2014 and 12\% in 2015.\\
    }
    \includegraphics[width=\textwidth]{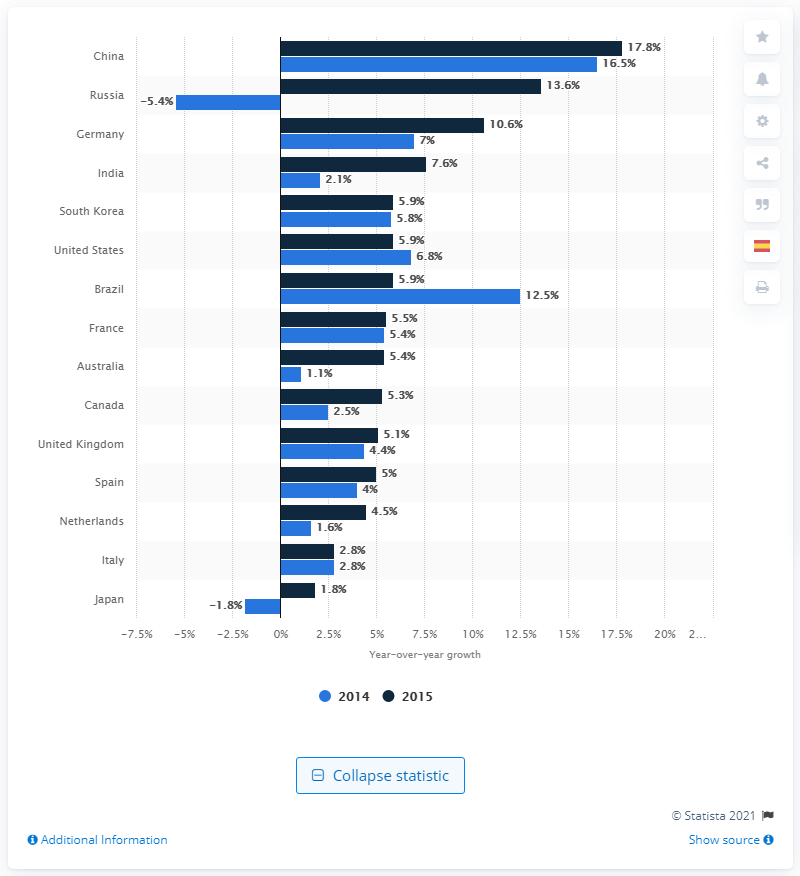}
    \label{fig:chart1}
\end{minipage}
\end{figure}

Our chart dataset is derived from \textit{Unichart}, \textit{Chart2Text}, and \textit{ChartQA}, covering real financial data, marketing trends, and varied visual styles. We focus on numerical and financial charts to support robust quantitative analysis. An example question-answer pair is shown in Figure~\ref{fig:chart1}.

\begin{figure*}
\centering
\begin{minipage}[t]{1\textwidth}
    \centering
    \caption{
        Example chart image~2.\\
        \textcolor{blue}{Instruction (Make Comparisons):}\\
        \textit{Q: How has the unemployment rate in the given chart changed from 1999 to 2020?}\\
        \textit{A:} The unemployment rate in the chart shows significant fluctuations between 1999 and 2020. Starting at 9.64\% in 1999, it slightly increased to 9.65\% in 2000. There was a sharp decline to 6.81\% in 2003 and further to 4.83\% in 2005. The rate then fluctuated, peaking at 5.46\% in 2009 before decreasing to 4.09\% in 2011. It rose again to 5.26\% in 2016 and slightly varied around 4.62\% in 2018. By 2020, the rate was at 4.81\%. Overall, the chart indicates a general downward trend in the unemployment rate over the two decades with occasional increases.\\
    }
    \includegraphics[width=0.8\textwidth]{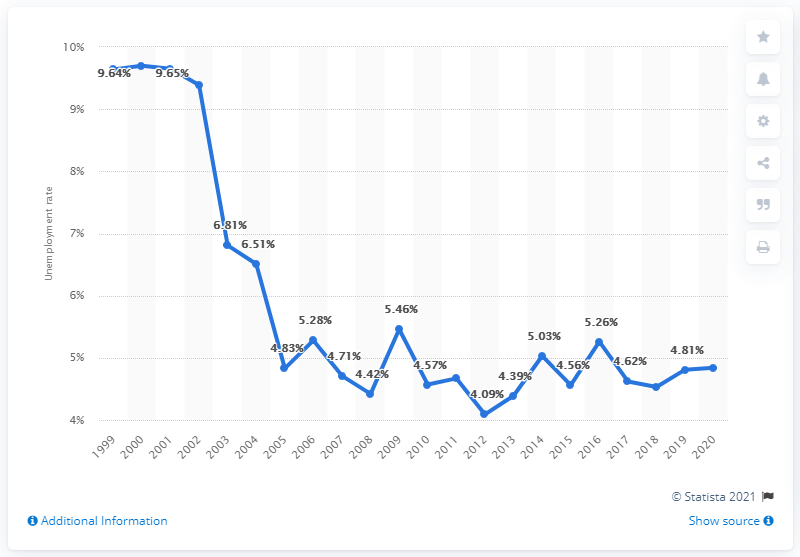}
    \label{fig:chart2}
\end{minipage}
\end{figure*}

During SFT, we develop seven chart-specific tasks (Appendix~\ref{appendix-tasks}) to enhance the model's ability to interpret financial charts and extract insights. Figure~\ref{fig:chart2} illustrates an example chart used in SFT data generation.

\clearpage
\fcolorbox{black}{gray!10}{
    \parbox{0.9\textwidth}{ 
You are a data analyst reviewing a table from a financial report. Your task is to understand the data and its location in the table.
Based on the dataset's table structure and content, interpret what the table represents.\\
\textbf{Cell Information:}\\
Each cell in the dataset is described with four main attributes:\\
- \textbf{bbox:} Bounding box coordinates indicating the position of the cell within the table.\\
- \textbf{tokens:} The actual content of the cell, which might include text or numerical data.\\
- \textbf{is\_header:} A boolean value indicating whether the cell is a header cell or not.\\
- \textbf{span:} Additional information about the cell's span, such as colspan and rowspan.\\
\textbf{Details of the cells:}\\
\textcolor{blue}{\{cells\_str\}}\\
\textbf{Tasks for Pretraining Data:}\\
- Examine the table's content to understand what information it conveys and how it is structured.\\
- Generate a descriptive question that encourages a deep dive into the table’s displayed data and its organizational framework.\\
- Avoid using HTML tags in your response.\\
- Ensure your response contains numerical data.\\
- Provide detailed information about the table's content and structure, highlighting specific values from the table (excluding headers).\\
- Provide spatial information about the cells in this table without including any bounding box information. For example, you can describe the column or row layout.\\
\textbf{Tasks for SFT Data:}\\
- Understand the content of the table, including the spatial position of each cell, numerical information, headers, etc.\\
- Based on the specific content of the table, propose a question related to \textcolor{blue}{\{selected\_task\}} for a deeper analysis of the table.\\
- If the table content is related to data, analyze the data and use the numerical values provided to support your analysis.\\
- If you don't know the answer to the question, you should further analyze the table to provide a more detailed examination of its content.\\
\textbf{Note:}\\
 You should not create or invent any numerical values; you should only analyze using the values provided in the table.
    }
}

\clearpage
\fcolorbox{black}{gray!10}{%
  \parbox{0.9\textwidth}{
    Based on the provided chart image, generate a detailed caption or description that thoroughly explains the chart's content, including its numerical data and organizational structure.\\
\\
    \textbf{Tasks for Pretraining Data:}\\
    - Examine the chart's content to understand what information it conveys and how it is structured.\\
    - Generate a comprehensive description or caption that covers all aspects of the chart’s numerical data and organizational framework.\\
    \\
\textbf{Random Questions for Pretraining Data:}
\begin{itemize}
    \item What key information does the chart present?
    \item Can you detail the main data points covered in the chart?
    \item What topics or categories are addressed within the chart?
    \item Identify the major types of data provided by the chart.
    \item What essential insights does the chart offer about its subject matter?
    \item What are the primary themes or elements highlighted in the chart?
    \item Describe the type of information that the chart provides.
    \item What are the main conclusions or facts presented in the chart?
    \item Outline the core pieces of information contained in the chart.
    \item Summarize the main subjects or topics included in the chart.
    \item What does the chart specifically focus on in its data presentation?
\end{itemize}
  }
}

\clearpage
\begin{center}
\label{fig:chart-prompt}
\fcolorbox{black}{gray!10}{
\parbox{1\textwidth}{
Your task is to analyze a financial chart represented in an image. Follow these detailed steps to understand the data and its location in the chart, and generate a high-quality SFT dataset.

\textbf{Tasks for SFT Data:}
\begin{enumerate}
    \item \textbf{Interpret the Chart's Content:}
    \begin{itemize}
        \item Identify the spatial position of each cell in the chart.
        \item Only extract and note the numerical information which is 100\% convincing to you from the chart.
        \item Identify and record headers and labels from the chart.
    \end{itemize}
    \item \textbf{Formulate a Question for Deeper Analysis:}
    \begin{itemize}
        \item Based on the specific content of the chart, propose a question related to \textcolor{blue}{\{selected\_task\}}.
    \end{itemize}
    \item \textbf{Analyze the Data:}
    \begin{itemize}
        \item If the chart content includes data, analyze it.
        \item Use the numerical values provided in the chart to support your analysis.
    \end{itemize}
    \item \textbf{Further Examination:}
    \begin{itemize}
        \item If you are unsure of the answer to the question, perform a deeper analysis of the table to provide a more detailed examination of its content.
    \end{itemize}
\end{enumerate}

\textbf{Note:} Do not create or invent any numerical values; only use the values provided in the table.

\textbf{Example Chart Analysis Workflow:}
\begin{enumerate}
    \item \textbf{Interpret the Chart's Content:}
    \begin{itemize}
        \item Examine the image to identify the layout of the chart, such as rows and columns.
        \item Record numerical values and their corresponding labels or headers.
    \end{itemize}
    \item \textbf{Formulate a Question for Deeper Analysis:}
    \begin{itemize}
        \item Example: ``What is the trend of quarterly sales over the past year?''
    \end{itemize}
    \item \textbf{Analyze the Data:}
    \begin{itemize}
        \item Example: ``The quarterly sales over the past year show an increasing trend, starting from \$50,000 in Q1 to \$70,000 in Q4. This indicates consistent growth in sales, with Q2 and Q3 showing moderate increases to \$55,000 and \$60,000 respectively.''
    \end{itemize}
    \item \textbf{Further Examination:}
    \begin{itemize}
        \item If further detail is required, delve deeper into the data and provide more specific insights.
    \end{itemize}
\end{enumerate}
}}
\end{center}

\clearpage

\clearpage

\clearpage
\section{Prompts for Generating Finance-Relevance Score}

\begin{center}
\fcolorbox{black}{gray!10}{
\parbox{1\columnwidth}{
You are a financial analyst specialized in evaluating the relevance of a conversation to the financial domain. Your goal is to assess how closely the conversation content pertains to financial topics.

\textbf{Conversation:}

Human: \textcolor{blue}{\{conversation[0][``value'']\}} 

GPT: \textcolor{blue}{\{conversation[1][``value'']\}}

\textbf{Tasks:}

You need to complete the following tasks:
- Review the conversation between the human and the GPT model, and determine the relevance of the conversation to financial matters.
- Provide a relevance score between 0 and 10, where 0 indicates no relevance and 10 indicates high relevance to finance.
- Ensure the result is formatted correctly in JSON for further analysis.

\textbf{Output Format:}

\{``relevance\_score": ``<Insert relevance score here>"\}
}}
\end{center}

\section{Experiments}
\subsection{Datasets used for different experiments}
\label{appendix-experiment-dataset}
Details of the dataset used for evaluation are listed in Table ~\ref{tab:zero-shot-dataset-info}, \ref{tab:few-shots-dataset-info}, \ref{inst_eval_datasets}, \ref{tab:multimodal-data-info}.

\begin{table}[t]
\centering
\scriptsize
\caption{Datasets used for zero-shot evaluation tasks. The dataset with * is newly created in this paper. Abbreviations: EM Accuracy means exact match accuracy. MCC means Matthews correlation coefficient.}
\label{tab:zero-shot-dataset-info}
\resizebox{\columnwidth}{!}{
\scalebox{0.8}{
\begin{tabular}{lllll}
\toprule
\textbf{Task}&\textbf{Dataset} & \textbf{Test Size} & \textbf{Metrics} & \textbf{License} \\ 
\midrule
Sentiment analysis&TSA~\citep{cortis2017semeval} & 561 & F1 & CC BY-NC-SA 4.0 \\
Classification&FOMC~\citep{shah2023trillion} & 496 & F1 & CC BY-NC 4.0 \\
Classification&FinArg-AUC~\citep{chen2023overview} & 969 & F1 & CC BY-NC-SA 4.0 \\
Classification&MA~\citep{yang2020generating} & 500 & F1 & Public \\
Classification&SC~\citep{mariko2020financial} & 8,630 & Entity F1 & CC BY 4.0 \\
Misinformation&FinFact~\citep{rangapur2023fin} & 3,369 & Weighted F1 & Public \\
Math&*MC & 50 & F1 & Public \\
Math&KnowledgeMath~\citep{zhao2023knowledgemath} & 1000 & EM Accuracy & MIT License \\
Math&DocMath-Eval~\citep{zhao2023docmath} & 3200 & EM Accuracy & MIT License \\
Credit scoring&German~\citep{misc_german_credit_data_144} & 1,000 & MCC & CC BY 4.0 \\
Credit scoring&Australian~\citep{misc_australian_credit_approval_143} & 690 & MCC & CC BY 4.0 \\
Credit scoring&LendingClub~\citep{feng2024empowering} & 2,690 & MCC & CC0 1.0 \\
Fraud detection&ccf~\citep{feng2024empowering} & 2,278 & MCC & DbCL v1.0 \\
Fraud detection&ccfraud~\citep{feng2024empowering} & 2,097 & MCC & Public \\
Distress&polish~\citep{feng2024empowering} & 1,736 & MCC & CC BY 4.0 \\
Distress&taiwan~\citep{feng2024empowering} & 1,364 & MCC & CC BY 4.0 \\
Claim analysis&ProtoSeguro~\citep{feng2024empowering} & 2,381 & MCC & Public \\
Claim analysis&travelinsurance~\citep{feng2024empowering} & 3,800 & MCC & ODbL v1.0 \\
Question answering&ConvFinQA~\citep{chen2022convfinqa} & 1,490 & EM Accuracy & MIT License \\
\bottomrule
\end{tabular}} }
\end{table}

\begin{table}[h]
\centering
\scriptsize
\caption{Datasets used for few-shot evaluation tasks.}
\label{tab:few-shots-dataset-info}
\resizebox{\columnwidth}{!}{%
\begin{tabular}{lllll}
\toprule
\textbf{Task}&\textbf{Dataset} & \textbf{Test Size} & \textbf{Metrics} & \textbf{License} \\ 
\midrule
Sentiment analysis&FPB~\citep{malo2014good} & 970 & F1 & CC BY-SA 3.0 \\
Sentiment analysis&FiQA-SA~\citep{maia2018www} & 235 & F1 & Public \\
Classification&Headlines~\citep{sinha2021impact} & 2,283 & Avg F1 & CC BY-SA 3.0 \\
Named entity recognition&NER~\citep{alvarado2015domain} & 980 & Entity F1 & CC BY-SA 3.0 \\
\bottomrule
\end{tabular}}
\end{table}

\begin{table}[h]
\renewcommand{\arraystretch}{1.2}
\caption{Datasets used for evaluating instruction fine-tuned models.}
\label{inst_eval_datasets}
\resizebox{\columnwidth}{!}{%
\scalebox{0.75}{
\begin{tabular}{lllll}
\hline
\textbf{Task}  
& \textbf{Dataset}    
& \textbf{Test size}
& \textbf{Metrics}        
& \textbf{License} 
\\ \hline
\multirow{4}{*}{\begin{tabular}[c]{@{}l@{}}Sentiment analysis \end{tabular}}
& FiQA-SA~\citep{maia2018www}   
& 235  
& \multirow{4}{*}{Accuracy}   
& Public        
\\
& FOMC~\citep{shah2023trillion}  
& 496   
&         
& CC BY-NC 4.0    
\\
& FPB~\citep{malo2014good}  
& 970  
&          
& CC BY-SA 3.0       
\\
& Headlines~\citep{sinha2021impact}
& 2,283    
&            
& CC BY-SA 3.0    
\\ \hline
\multirow{2}{*}{\begin{tabular}[c]{@{}l@{}}Named entity recognition \end{tabular}} 
& Finer-Ord~\citep{shah2023finer}
& 1,080  
& \multirow{2}{*}{Entity-F1} 
& CC BY-NC 4.0     
\\
& NER~\citep{alvarado2015domain}   
& 980  
&          
& CC BY-SA 3.0       
\\ \hline
\multirow{2}{*}{\begin{tabular}[c]{@{}l@{}}Number understanding \end{tabular}}    
& ConvFinQA~\citep{chen2022convfinqa} 
& 1,490   
& \multirow{2}{*}{EM Accuracy}
& MIT License        
\\
& FinQA~\citep{chen2021finqa}   
& 1,147  
&          
& MIT License      
\\ \hline
\multirow{2}{*}{\begin{tabular}[c]{@{}l@{}}Text summarization\end{tabular}}    
& ECTSUM~\citep{mukherjee2022ectsum}    
& 495   
& \multirow{2}{*}{Rouge-score} 
& Public       
\\
& EDTSUM~\citep{zhou2021trade} 
& 2,000    
&         
& Public     
\\ \hline
\multirow{3}{*}{\begin{tabular}[c]{@{}l@{}}Stock movement prediction \end{tabular}}
& ACL18~\citep{xu2018stock}   
& 3.720
& \multirow{3}{*}{Accuracy}    
& MIT License        
\\
& BigData22~\citep{soun2022accurate}
& 1,470  
&          
& Public      
\\
& CIKM18~\citep{wu2018hybrid}    
& 1,140    
&           
& Public      
\\ \hline
\multirow{2}{*}{Credit scoring}   
& Australian~\citep{misc_australian_credit_approval_143} 
& 690     
& \multirow{2}{*}{Accuracy}  
& CC BY 4.0       
\\
& German~\citep{misc_german_credit_data_144} 
& 1,000 
&        
& CC BY 4.0      
\\ \hline
\end{tabular}}}
\end{table}

\begin{table}[!h]
\centering
\small
\caption{Datasets used for multimodal evaluation tasks. }
\label{tab:multimodal-data-info}
\resizebox{\columnwidth}{!}{%
\begin{tabular}{llll}
\toprule
\textbf{Dataset} & \textbf{Test Size} & \textbf{Metrics} & \textbf{License} \\ 
\midrule
MMMU~\citep{yue2023mmmu} & 10,500 & Accuracy & Public  \\
MMMU-Business~\citep{yue2023mmmu} & 1,428 & Accuracy & Public \\
ChartBench & 350 & Accuracy & our data \\
TableBench & 450 & Accuracy & our data \\
\bottomrule
\end{tabular}}
\end{table}

\subsection{Details of Agent evaluation results}
\label{agent_results}
Results of different models in agent evaluation setting is lised in \ref{agent_results}.

\begin{table}[!h]
\renewcommand{\arraystretch}{0.9} 
\setlength{\tabcolsep}{4pt}      
\scriptsize
\centering
\renewcommand{\arraystretch}{1.3}
\caption{Trading performance of FinLLaMA and baseline models.}
\label{tab:trading-performance}
\resizebox{\columnwidth}{!}{%
\scalebox{0.9}{
\begin{tabular}{llcccccc}
\hline
\textbf{Metric} & \textbf{Asset} & \textbf{Buy \& Hold} & \textbf{LLaMA3.1-8B} & \textbf{LLaMA3-8B} & \textbf{FinLLaMA} \\ \hline
Cumulative Return & TSLA & -0.1633 & -0.1829 & -0.1552 & \textbf{0.5573} \\ 
& COIN & -0.0562 & -0.3515 & -0.0319 & \textbf{0.1743} \\ 
& GOOG & -0.0562 & 0.1651 & 0.1631 & \textbf{0.1098} \\ 
& NIO & -0.3530 & -0.2677 & -0.3526 & \textbf{0.4645} \\ 
& Overall & -0.1571 & -0.1592 & -0.0942 & \textbf{0.3265} \\ \hline
Sharpe Ratio & TSLA & -0.4769 & -0.5557 & -0.8486 & \textbf{2.4532} \\ 
& COIN & -0.1041 & -0.7911 & -0.0666 & \textbf{0.5778} \\ 
& GOOG & -0.1041 & 0.9464 & 1.0128 & \textbf{0.6927} \\ 
& NIO & -0.8678 & -0.6951 & -1.0313 & \textbf{1.9113} \\ 
& Overall & -0.3823 & -0.2739 & -0.2334 & \textbf{1.4088} \\ \hline
Normalized Sharpe Ratio & TSLA & 42.05 & 40.72 & 35.86 & \textbf{90.89} \\ 
& COIN & 48.27 & 36.82 & 48.89 & \textbf{59.63} \\ 
& GOOG & 48.27 & 82.44 & 83.88 & \textbf{61.55} \\ 
& NIO & 35.54 & 38.41 & 32.81 & \textbf{81.86} \\ 
& Overall & 43.63 & 45.43 & 45.92 & \textbf{73.48} \\ \hline
Annual Volatility & TSLA & 0.7015 & 0.6741 & 0.3748 & \textbf{0.4654} \\ 
& COIN & 1.1047 & 0.9103 & 0.9823 & \textbf{0.6181} \\ 
& GOOG & 1.1047 & 0.3574 & 0.3300 & \textbf{0.3249} \\ 
& NIO & 0.8332 & 0.7891 & 0.7005 & \textbf{0.4979} \\ 
& Overall & 0.9360 & 0.6827 & 0.5969 & \textbf{0.4766} \\ \hline
Max Drawdown & TSLA & 0.5532 & 0.5248 & 0.2662 & \textbf{0.1883} \\ 
& COIN & 0.6019 & 0.5449 & 0.4056 & \textbf{0.4142} \\ 
& GOOG & 0.6019 & 0.1983 & 0.2092 & \textbf{0.1741} \\ 
& NIO & 0.4498 & 0.4213 & 0.5172 & \textbf{0.3008} \\ 
& Overall & 0.5517 & 0.4223 & 0.3496 & \textbf{0.2693} \\ 
\hline
\end{tabular}}}
\end{table}

\subsection{Multimodal Task Categories}
\label{appendix-tasks}
We chose these seven categories for our evaluations because they represent the most critical and common tasks in financial analysis. By focusing on these areas, we ensure a comprehensive and thorough assessment of our model's capabilities in handling financial data.

\begin{itemize}
   \item \textbf{Make Comparisons}: This category involves comparing different financial metrics or data points across various time periods, companies, or financial instruments. For example, comparing quarterly revenues of different companies to determine market performance trends. This is crucial for financial analysts who need to benchmark performance and identify trends over time. Accurate comparisons help in making informed decisions about investments, cost management, and strategic planning.
    
   \item \textbf{Find Correlations}: This involves identifying relationships between different financial variables. For example, determining if there's a correlation between interest rates and stock prices, which can help in predictive financial modeling. 
Understanding correlations is essential for risk management and portfolio diversification, as it allows analysts to predict how changes in one variable might affect another.
    
   \item \textbf{Data Retrieval}: This category focuses on extracting specific data points from financial tables or charts. For example, retrieving the net income values from an annual financial report for analysis. Efficient data retrieval is fundamental for compiling reports, conducting audits, and performing detailed financial analysis. It ensures that all necessary data can be quickly accessed and utilized.
    
   \item \textbf{Find Extremum}: This involves identifying the maximum or minimum values within financial datasets. For example, finding the highest stock price over a given period or the lowest expense in a budget report. Identifying extremum points helps in spotting significant events or trends that might require further investigation or immediate action. This is particularly useful in scenarios like peak revenue analysis or cost-cutting strategies.
    
   \item \textbf{Find Clusters}: This category entails grouping financial data into clusters based on similarities. For example, clustering companies based on similar financial performance indicators like revenue, profit margins, and market share. Clustering helps in market segmentation, identifying peer groups, and understanding competitive positioning. It is valuable for comparative analysis and strategic planning.
    
   \item \textbf{Characterize Distributions}: This involves describing the distribution of financial data points. For example, analyzing the distribution of daily returns of a stock to understand its volatility and risk. Characterizing distributions aids in risk assessment, financial forecasting, and identifying patterns that could influence decision-making processes. It provides a statistical foundation for understanding variability and risk.
    
   \item \textbf{Find Anomalies}: This focuses on detecting outliers or unusual patterns in financial data. For example, identifying unexpected spikes in expenses that could indicate fraud or errors in financial reports. Detecting anomalies is crucial for maintaining the integrity of financial data, preventing fraud, and ensuring accurate financial reporting. It helps in early detection of issues that might otherwise go unnoticed.
\end{itemize}
\subsection{Details of Continual Pretrained Model Evaluation}

\subsubsection{Descriptions of Zero-shot Evaluation Tasks}
\label{append:basemodel_eval_task_details}
\begin{itemize}
\item \textbf{Sentiment
analysis} focuses on extracting sentiment information (positive, negative, or neutral) from financial
texts, using the TSA dataset.
\item \textbf{Classfication}: 1) Hawkish-Dovish classification aims to classify sentences from monetary policy texts
as ’hawkish’ or ’dovish’ focusing on the nuanced language and economic implications of financial
texts, using the FOMC dataset. 2) Argument unit classification categorizes sentences as claims or premises, using the FinArg AUC dataset. 3) Deal completeness classification predicts if mergers and acquisitions events are "completed" or remain "rumors" based on news and tweets, employing the MA dataset.
\item \textbf{Causal Classification} discerns whether sentences from financial news and SEC filings convey causality, using the SC dataset.
\item \textbf{Misinformation} detection is formulated as a three-classification task, verifying financial misinformation (True/False/Not Enough Information). The input is textual claim information. The aim is to let the model deliver accurate results, which requires LLMs to identify fraudulent financial content and verify the claim's authenticity.
\item \textbf{Mathematical Computation} is structured as a generation task, specifically designed to compute financial metrics based solely on questions about company financial statements. The input for this task consists of 50 questions, each focusing on different financial metrics such as revenue, turnover, and other measurable outcomes. The objective is for the model to generate accurate financial analyses, such as capital expenditures or financial ratios, from the information presented in the questions alone, without direct access to the financial sheets. This task assesses the model's ability to infer and calculate key financial indicators crucial for evaluating the financial health and performance of the company.
\item \textbf{KnowledgeMath} is formulated as a math word problem-solving task, predicting the value of the final answer. The input is a math word problem in finance domains, the aim is to let the model perform math reasoning to predict the final answer of the math word problem.
\item \textbf{DocMath-Eval} is formulated as the document question answering task, predicting the value of the final answer. The input comprises a financial document and a question, the aim is to let the model perform information extraction and math reasoning to predict the final answer of the question.
\item \textbf{Credit Scoring} is a vital process employed by financial institutions to evaluate a borrower's creditworthiness. It assesses financial information provided in loan applications to determine eligibility, interest rates, and loan terms to predict credit risk.
\item \textbf{Fraud Detection} is a task closely aligned with credit scoring, focusing on identifying genuine versus fraudulent loan applications. This process is essential for safeguarding financial systems and shielding institutions from financial losses. The datasets of this task are often imbalanced, a characteristic common to fraud detection, with genuine fraud cases constituting a small fraction of total applications.
\item \textbf{Financial Distress Identification} aims to predict the likelihood of a company experiencing bankruptcy, leveraging publicly accessible data. This process is crucial for stakeholders to assess the financial health and stability of a company.
\item \textbf{Claim Analysis} is a critical task for insurance companies, involving the analysis of claims to detect fraudulent activity. Fraudulent claims are illegitimate attempts to obtain payment under false pretenses, while legitimate claims represent valid requests for payment due to losses covered by an insurance policy. This distinction is vital for preventing financial losses due to fraud and ensuring that only rightful claims are reimbursed. Most datasets of this task are often imbalanced, meaning that fraudulent claims are significantly less frequent than legitimate ones, a common scenario in real-world insurance claim analysis.
\item \textbf{Question Answering} focuses on answering financial questions based on the provided information. We use the ConFinQA dataset, which includes multi-turn question-and-answer pairs over earnings reports.
\end{itemize}
\subsubsection{Descriptions of Few-shot Evaluation Tasks}
\label{append:few_shot_task_desrip}
\begin{figure*}[t]
\centering
\includegraphics[width=0.99\linewidth]{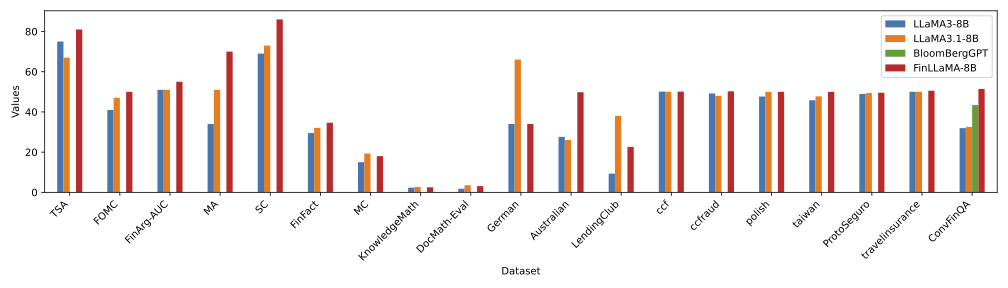}
\caption{Zero-shot performance of FinLLaMA and baseline models.}
\label{fig:zero-shot-performance}
\end{figure*}

\begin{figure}[t]
\centering
\includegraphics[width=0.9\columnwidth]{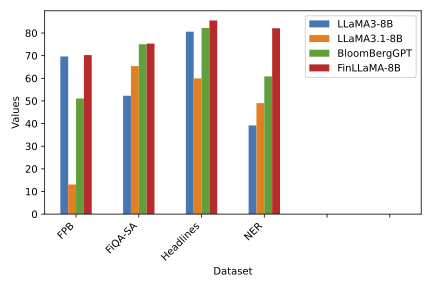}
\caption{Few-shot performance of FinLLaMA and baseline models.}
\label{fig:few-shot-performance}
\end{figure}

In our few-shot evaluations, we use three financial NLP tasks:
\begin{itemize}
    \item \textbf{Sentiment Analysis}: Extracting sentiment information from financial texts using the FPB and FiQA-SA datasets, which focus on determining sentiment polarity (positive, negative, or neutral) in financial sentences.
     \item \textbf{Classification}: Evaluating the model's capability to classify financial texts. The Headlines dataset is used to classify news headlines related to financial events.
     \item \textbf{Named Entity Recognition}: Extracting entities such as persons, organizations, and locations from financial texts. We use the NER dataset with manually annotated four entities for three financial agreements.
 \end{itemize}
\label{append:icl_prompt}
For detailed prompts for evaluation datasets, please see Table \ref{tab:prompt_0_shot} for 0-shot, and Table \ref{tab:prompt_icl} for few-shots. The results are listed in Figure \ref{fig:zero-shot-performance} and Figure \ref{fig:few-shot-performance}.

\begin{table*}[htb!]
\centering
\scriptsize
\caption{0-shot task datasets prompt overview.}
\scalebox{0.95}{
\begin{tabular}{ll}
\toprule
\textbf{Data} 
& \textbf{Prompt} \\
\midrule

\textcolor{black}{TSA} & \makecell[l]{``Given the following financial text, return a sentiment score for Ashtead as a floating-point number \\ranging from -1 (indicating a very negative or bearish sentiment) to 1 (indicating a very positive or bullish sentiment), \\with 0 designating neutral sentiment. Return only the numerical score first, \\follow it with a brief reasoning behind your score."} 
\\ \midrule

\textcolor{black}{FOMC} & \makecell[l]{\textcolor{black}{``Examine the excerpt from a central bank's release below. Classify it as HAWKISH if it advocates for a tightening} \\of monetary policy, DOVISH if it suggests an easing of monetary policy, or NEUTRAL if the stance is unbiased.\\Your response should return only HAWKISH, DOVISH, or NEUTRAL."}
\\ \midrule

\textcolor{black}{FinArg - ACC} & \makecell[l]{``Analyze sentences from earnings conference calls and identify \\their argumentative function. \\Each sentence is either a premise, offering evidence or reasoning, or a claim, \\asserting a conclusion or viewpoint. Return only premise or claim."} 
\\ \midrule

\textcolor{black}{MA} & \makecell[l]{``In this task, you will be given Mergers and Acquisitions news articles or tweets. \\Your task is to classify each article or tweet based on whether the mentioned deal was completed or remained a rumour. \\Your response should be a single word - either 'complete' or 'rumour' - \\representing the outcome of the deal mentioned in the provided text."} 
\\ \midrule

\textcolor{black}{SC} & \makecell[l]{``In this task, you are provided with sentences extracted from financial news and SEC data. \\Your goal is to classify each sentence into either 'causal' or 'noise' based on whether or not it indicates a causal relationship between financial events. \\Please return only the category 'causal' or 'noise'."} 
\\ \midrule

FinFact & \makecell[l]{`` Determine if the following claim is 0. True or 1. False or 2. NEI (Not Enough Information). Please directly answer and do not explain.
Claim: "} 
\\ \midrule

MC & \makecell[l]{``
Please answer following questions: "} 
\\ \midrule

KnowledgeMath & \makecell[l]{`` You are a financial expert, you are supposed to answer the given question. \\You need to first think through the problem step by step, documenting each necessary step. \\Then you are required to conclude your response with the final answer in your last sentence as \\“Therefore, the answer is \{final answer\}”. The final answer should be a numeric value.\\
Question: \textcolor{blue}{\{question\}}
\\Let’s think step by step to answer the given question. "}
\\ \midrule

DocMath-Eval & \makecell[l]{`` You are a financial expert, you are supposed to answer the given question. \\You need to first think through the problem step by step, documenting each necessary step. \\Then you are required to conclude your response with the final answer in your last sentence as \\“Therefore, the answer is \{final answer\}”. The final answer should be a numeric value.\\
\textcolor{blue}{\{Document context\}}
\\Let’s think step by step to answer the given question. "}
\\ \midrule
        
\textcolor{black}{German} & \makecell[l]{\textcolor{black}{``Assess the creditworthiness of a customer using the following table attributes for financial status. Respond with either} \\ \textcolor{black}{'good' or 'bad'. And the table attributes including 13 categorical attributes and 7 numerical attributes are as follows:"}}
\\ \midrule

\textcolor{black}{Australian} & \makecell[l]{``Assess the creditworthiness of a customer using the following table attributes for financial status. Respond with either \\'good' or 'bad'. And the table attributes including 13 categorical attributes \\ and 7 numerical attributes and values have been changed to meaningless symbols to protect confidentiality of the data. :"}\\ \midrule
 
\textcolor{black}{LendingClub} & \makecell[l]{``Assess the client's loan status based on the following loan records from Lending Club. \\Respond with only 'good' or 'bad', and do not provide any additional information. \\For instance, 'The client has a stable income, no previous debts, and owns a property.' should be classified as 'good'."} 
\\ \midrule

\textcolor{black}{ccf} & \makecell[l]{``Detect the credit card fraud using the following financial table attributes. \\Respond with only 'yes' or 'no', and do not provide any additional information. \\Therein, the data contains 28 numerical input variables V1, V2, ..., \\and V28 which are the result of a PCA transformation and 1 input variable Amount which has not been transformed with PCA. \\The feature 'Amount' is the transaction Amount, this feature can be used for example-dependant cost-sensitive learning. \\For instance, 'The client has attributes:\textcolor{blue}{\{category\}}"} 
\\ \midrule

\textcolor{black}{ccfraud} & \makecell[l]{``Detect the credit card fraud with the following financial profile. \\Respond with only 'good' or 'bad', and do not provide any additional information. For instance, \\'The client is a female, the state number is 25, the number of cards is 1, the credit balance is 7000, \\the number of transactions is 16, the number of international transactions is 0, \\the credit limit is 6.' should be classified as 'good'."} 
\\ \midrule

\textcolor{black}{polish} & \makecell[l]{``Predict whether the company will face bankruptcy based on the financial profile attributes provided in the following text. \\Respond with only 'no' or 'yes', and do not provide any additional information."} 
\\ \midrule

\textcolor{black}{taiwan} & \makecell[l]{``Predict whether the company will face bankruptcy based on the financial profile attributes provided in the following text. \\Respond with only 'no' or 'yes', and do not provide any additional information."} 
\\ \midrule

\textcolor{black}{Porto-Seguro} & \makecell[l]{``Identify whether or not to files a claim for the auto insurance policy holder using the following table attributes about individual financial profile. \\Respond with only 'yes' or 'no', and do not provide any additional information.\\ And the table attributes that belong to similar groupings are tagged as such in the feature names (e.g., ind, reg, car, calc). \\In addition, feature names include the postfix bin to indicate binary features and cat to indicate categorical features. \\Features without these designations are either continuous or ordinal. \\Values of -1 indicate that the feature was missing from the observation."} 
\\ \midrule

\textcolor{black}{travelinsurace} & \makecell[l]{``Identify the claim status of insurance companies using the following table attributes for travel insurance status. \\Respond with only 'yes' or 'no', and do not provide any additional information.\\ And the table attributes including 5 categorical attributes and 4 numerical attributes are as follows:\textcolor{blue}{\{category\}}"} \\
        
\bottomrule
\end{tabular}}
\label{tab:prompt_0_shot}
\end{table*}

\begin{table*}[htb!]
\centering
\scriptsize
\caption{Few-shot task datasets prompt overview.}
\scalebox{0.95}{
\begin{tabular}{llp{4.1cm}}
\toprule
\textbf{Data} 
& \textbf{Prompt} \\
\midrule
FPB & \makecell[l]{``Analyze the sentiment of this statement extracted from a financial news article. \\Provide your answer as either negative, positive, or neutral. \\For instance, 'The company's stocks plummeted following the scandal.' would be classified as negative."} &
\\ \midrule

FiQA-SA & \makecell[l]{``What is the sentiment of the following financial \textcolor{blue}{\{category\}}:\\Positive, Negative, or Neutral?"} 
\\ \midrule

\textcolor{black}{Headlines} & \makecell[l]{``Consider whether the headline mentions the price of gold.\\Is there a Price or Not in the gold commodity market indicated in the news headline?\\Please answer Yes or No."} 
\\ \midrule

\textcolor{black}{NER} & \makecell[l]{``In the sentences extracted from financial agreements in U.S. SEC filings,\\identify the named entities that represent a person ('PER'), an organization ('ORG'),\\or a location ('LOC'). The required answer format is: 'entity name, entity type'.\\For instance, in 'Elon Musk, CEO of SpaceX, announced the launch from Cape Canaveral.',\\the entities would be: 'Elon Musk, PER; SpaceX, ORG; Cape Canaveral, LOC'"}
\\ \bottomrule
\end{tabular}}
\label{tab:prompt_icl}
\end{table*}

\label{appendix:zero-shot}

\label{appendix:few-shots}

\subsection{Prompts for Multimodal Benchmark Evaluation}

In our multimodal benchmark evaluation, we introduce two main stages. First, we generate answers from LLMs using three different templates of prompts. Second, we utilize gpt-4o-mini to extract the correct JSON format from the LLM's response.

The results in Table \ref{tab:multimodal_results} display only the highest scores attained by any of the three prompts for each evaluation metric. This approach highlights the optimal performance our model can achieve, leveraging the strengths of each prompt. By doing so, we provide a clear and concise summary of our model's capabilities, illustrating its versatility and robustness across different evaluation scenarios.

\subsection{Prompts for Generating LLM Answers}
In our evaluation, we utilized three distinct prompts to assess performance comprehensively:\\

$Prompt~0$: Our custom-designed prompt is tailored specifically for the unique characteristics of our dataset and objectives.

$Prompt~1$: The standard prompt corresponds to the MMbench benchmark, allowing for a standardized comparison with existing models.

$Prompt~2$: The prompt associated with FinTral~\citep{bhatia2024fintral} , ensuring relevance and applicability in financial domain-specific tasks.\\

The rationale behind employing these three prompts is to capture a broad spectrum of scenarios and requirements.

By including our customized prompt ($Prompt~0$), we can fine-tune our model to the specific nuances of our dataset. The inclusion of MMbench ($Prompt~1$) provides a standardized baseline for comparison, facilitating a fair assessment against other models in the field. Lastly, FinTral's prompt ($Prompt~2$) ensures that the evaluation encompasses domain-specific challenges, particularly in the financial sector.

\subsubsection{$Prompt~0$}

\begin{center}
\fcolorbox{black}{gray!10}{\parbox{.95\linewidth}{
\textless{}image\textgreater{}\textbackslash{}n Given the following multiple-choice financial question and the image that contains relevant information, generate the correct answer based on the image content:\\

\textbf{Question:} \textcolor{blue}{\{question\}}\\

\textbf{The output format should be:}\\
\{``answer": ``<Insert the correct option here>"\}
}}
\end{center}

\subsubsection{$Prompt~1$}

\begin{center}
\fcolorbox{black}{gray!10}{\parbox{.95\linewidth}{
\textless{}image\textgreater{}\textbackslash{}n \textcolor{blue}{\{question\}} Answer with the option's letter from the given choices directly.}}
\end{center}

\subsubsection{$Prompt~2$}

\begin{center}
\fcolorbox{black}{gray!10}{\parbox{.95\linewidth}{
You are a financial expert specializing in the nuanced analysis of financial statements and a wide array of data-driven financial tasks. For each prompt you are given, think step by step. Sometimes, you must extract relevant information to proceed with the problem. \\

\textbf{Instructions: }If any options are specified, ensure that your answer is one of the options specified.\\
Do not explain why you think the answer is correct. \\

\textbf{Context:}\textless{}image\textgreater{}\textbackslash{}n \\
\textbf{Question:}\textcolor{blue}{\{question\}}\\
\textbf{Answer:}

}}
\end{center}

\subsection{Example of LLM Answers}

\textbf{Question: }How does the stock index value on January 15, 2023, compare to the value on January 30, 2023?A. January 15 is lower B. January 15 is equal C. January 30 is higher D. January 15 is higher\\

Answer of $Prompt~0$:\textbf{:} \{``answer": ``A. January 15 is lower"\}

Answer of $Prompt~1$:A

Answer of $Prompt~2$:Identify Key Dates and Values: The graph shows a downward trend from early January. On January 15 (indicated by ``2023-01-17"), the stock index was around 4000 points. By January 30 (indicated as ``2023-01-29"), the stock index had decreased significantly below 3000 points at approximately 3300 points.

Compare Values: As per the data: Value on January 15 = ~4000; Value on January 30 = ~3300

Conclusion Based on Comparison: Since 4000 > 3300, it's clear that the stock index value on January 15, 2023, is higher compared to the value on January 30, 2023.

Therefore, based on this analysis:Answer: D. January 15 is higher

This conclusion aligns with option D because the values indicate an increase over time before reaching their peak again later into the month.

\begin{figure}[h]
\caption{Example input.}
\label{fig:example}
    \centering
        \includegraphics[width=0.85\linewidth]{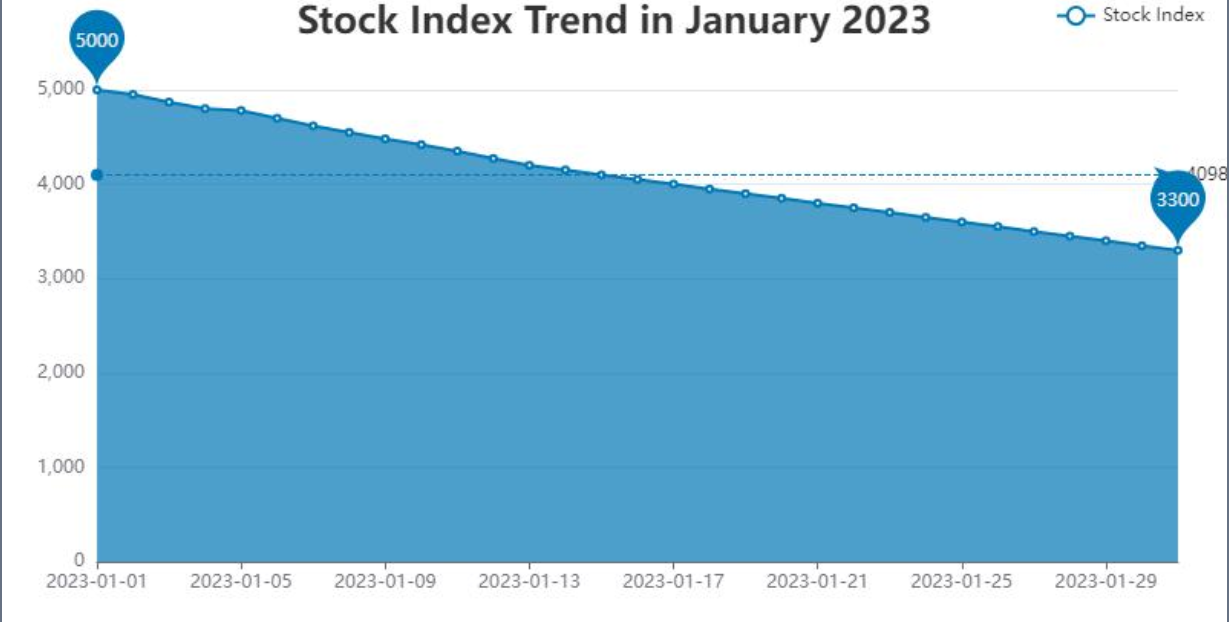}
\end{figure}

\subsection{Prompts for Evaluating LLM Answers}

For each prompt, we conducted separate evaluations to assess the performance of our LLM. To ensure the consistency and relevance of the LLM's generated answers, we employed GPT-4o-mini to filter and determine the accuracy of the outputs for each prompt individually. This filtering process helps in aligning the generated answers with the intended prompt format and maintaining output quality. Here is our prompt.

\begin{center}
\fcolorbox{black}{gray!10}{\parbox{.95\linewidth}{You are an AI model tasked with answering a multiple-choice question based on the provided additional information. Please answer the multiple-choice question based barely on the provided additional information. \\

\textbf{Question and options:} \textcolor{blue}{\{query\}}\\

\textbf{Additional Information:} \textcolor{blue}{\{LLM's output\}}\\

Please analyze the question and options carefully and provide the answer in the following JSON format:
\{``answer": ``<Insert the correct option here (A, B, C, or D)>"\}\\

\textbf{Note:}\\
- Remember that you should only give me the JSON format without any additional information.

- You should answer the question with only A, B, C, or D, not the full answer text.

- If the additional information tells the answer, then you should follow his options.

- If you don't know the answer, the output should be:\{``answer": ``E"\}
 }}
\end{center}

\section{Details for Trading in the Decision Making Task}
\label{trading-details}
In this task, at each time step, the LLM receives a feed of memories retrieved from the memory module. Based on this information, the LLM must make an investment decision—choosing to buy, sell, or hold—while providing its reasoning for the decision and specifying the index of the supporting information within the memory module. This process evaluates the model's ability to analyze financial data, make informed decisions, and justify its choices by referencing specific pieces of stored information in a dynamic trading environment.\newline
\textbf{Data}: We evaluated the model's performance on the Decision Making task using the following datasets:
\begin{itemize}
\item \textbf{OHLCV data}: Open-High-Low-Close prices and trading volume data for COIN, GOOG, NIO, and TSLA, obtained from Yahoo Finance.
\item \textbf{News data}: Collected from Alpaca News API for COIN, GOOG, NIO, and TSLA.
\item \textbf{Form 10-Q data}: Extracted from SEC EDGAR for COIN, GOOG, and TSLA.
\item \textbf{Form 10-K data}: Retrieved from SEC EDGAR for COIN, GOOG, and TSLA.
\end{itemize}
Figure~\ref{fig:trading-prompt1} and~\ref{fig:trading-prompt2} present our prompts for trading in the Decision Making task.

\begin{figure*}[ht!]
\begin{tcolorbox}
    \begin{tcolorbox}[colback=white, colframe=lightgrey, title=Initialize Profile, boxsep=2pt, top=2pt, bottom=2pt]
    \small
    \raggedright
    \textbf{\textcolor{lightgrey}{1. Operations:}} \\
    - Provide a performance overview of the trading stock based on available data.\\
    - Set up the risk inclination as the key character of the trading agent.\\
    \textbf{\textcolor{lightgrey}{2. Range:}} Financial information such as the financial sectors, historical performance, and previous stock trends of the trading stock.\\
    \textbf{\textcolor{lightgrey}{3. Prompts:}} You are an experienced trading manager and investment firm. Your task is to make informed decisions on the given stock based on the provided information.\\
    \textcolor{red}{Under Self-Adaptive Risk Character Setting:} When historical momentum is positive, you are a risk-seeking investor. But when historical momentum is negative, you are a risk-averse investor.\\
    \textbf{\textcolor{lightgrey}{4. General background setting:}} \\
    You have accumulated a lot of information about the following sectors, so you are especially good at trading them: 1)Electric Vehicles (Automotive Sector). 2) Energy Generation and Storage...From year 2021 to 2022 September, Tesla's continued growth and solid financial performance over the defined period ...\\
    \end{tcolorbox}
    \begin{tcolorbox}[colback=white, colframe=lightblue, title=Summarize, boxsep=2pt, top=2pt, bottom=2pt]
    \small
    \raggedright
    \textbf{\textcolor{lightblue}{1. Operations:}} \\
    - Summarize different types of input information. \\
    - Distribute them to corresponding layers of the long-term memory database.\\
    \textbf{\textcolor{lightblue}{2. Range:}} Daily market news, Long Documents such as company 10-K and 10-Q reports \\
    \textbf{\textcolor{lightblue}{3. Prompts:}}\\
    - (1). Summarize the contents: Summarize the following documents into {1000} words.\\
    - (2). Comprehend the investment sentiment of news insights: The positive, neutral and negative scores are for understanding the investment sentiments, opinions, or emotions. For example, positive news about a company can lift investor sentiment, encouraging more buying activity, which in turn can push stock prices higher...\\
    \textbf{\textcolor{lightblue}{4. Outputs:}} \\
    (1). \textbf{To Shallow Memory Layer:} \\
    - [News \textcolor{darkblue}{(ID: 261)}] Here's How Much You Would Have Made Owning Tesla Stock In The Last 10 Years Tesla (NASDAQ:TSLA) has outperformed the market over the past 10 years by 50.69\% on an annualized basis producing an average annual return of 60.76\%. Currently, Tesla has a market capitalization of \$683.54 billion.... The sentiment is \textcolor{red}{\{positive\}}. \\
    - [News \textcolor{darkblue}{(ID: 278)}] Tesla Q3 Earnings Are Imminent. Can Nio Foreshadow What's To Come? What To Know Before The Print Tesla Inc (NASDAQ: TSLA) shares were trading down slightly Wednesday afternoon ahead of the automaker\'s third-quarter report, but the stock is up 6\% over the last five sessions... The sentiment is \textcolor{red}{\{positive\}}.\\
    - ... \\
    (2). \textbf{To Intermediate Memory Layer:} \\ 
    - [Form 10-Q \textcolor{darkblue}{(ID: 222)}] Tesla Q3 2022 revenues were \$21.5 billion, up 56\% year-over-year. Automotive sales revenue grew 56\% to \$17.8 billion driven by higher Model 3/Y and Model S/X deliveries. Gross automotive margin declined to 27.9\% due to cost inflation and factory ramps. Net income was \$3.3 billion, up 102\% year-over-year. Positive free cash flow was \$6.1 billion...  \\
    - [News \textcolor{darkblue}{(ID: 275)}] Tesla Q3 Earnings Highlights: Record Revenue, Operating Margin And Free Cash Flow, Tesla Semi Deliveries Coming In December Electric vehicle leader Tesla Inc (NASDAQ: TSLA) reported third-quarter financial results after market close Wednesday...The sentiment is \textcolor{red}{\{neutral\}}.  \\
    - [News \textcolor{darkblue}{(ID: 274)}] Tesla Preps For 2023 Cybertruck Launch, Will Make Battery Packs In California The Cybertruck is one of Tesla Inc. (NASDAQ: TSLA) most hotly anticipated, but also most delayed, products.
    - ...The sentiment is \textcolor{red}{\{negative\}}.\\
    (3). \textbf{To Deep Memory Layer:} \\ 
    - [News \textcolor{darkblue}{(ID: 161)}] Tesla Whale Trades Spotted A whale with a lot of money to spend has taken a noticeably bearish stance on Tesla. Looking at the options history for Tesla (NASDAQ:TSLA) we detected 477 strange trades. The sentiment is \textcolor{red}{\{positive\}}.\\
    - [Self-reflection \textcolor{darkblue}{(ID: 226)}] Given the short-term positive news score in the market for TSLA and a positive cumulative return, there is a high probability of continued growth in the short term. However, investor should be aware of potential threats in the mid-term market with competitors like General Motors, and Nio...
    \end{tcolorbox}
\end{tcolorbox}
\end{figure*}

\begin{figure*}[ht!]
\begin{tcolorbox}
    \begin{tcolorbox}[colback=white, colframe=lightpurple, title=Observe, boxsep=2pt, top=2pt, bottom=2pt]
    \small
    \raggedright
    \textbf{\textcolor{lightpurple}{1. Operations:}} Access and interpret market indicators such as current stock prices and historical momentum data. \\
    \textbf{\textcolor{lightpurple}{2. Range:}} Stock's daily adjusted closing price, historical momentum in the past $k$ days ($k=3$ in this case), etc. \\
    \textbf{\textcolor{lightpurple}{3. Prompts:}}\\
    - The information below provides a summary of stock price fluctuations over the previous few days, which is the "momentum" of a stock. It reflects the trend of a stock. Momentum is based on the idea that securities that have performed well in the past will continue to perform well, and conversely, securities that have performed poorly will continue to perform poorly.\\
    \textbf{\textcolor{lightpurple}{4. Outputs:}}\\
    - (1). The daily adjusted closing price of TSLA on \textcolor{red}{\{2022-10-25\}} is \textcolor{red}{\{\$222.42\}}.\\
    - (2). \textbf{Train:} On \textcolor{red}{\{2022-10-25\}}, the momentum of TSLA, indicated by the price difference between the current and the next trading \\
    \quad \quad \; day, is \textcolor{red}{\{\$2.22\}}.\\
    \quad \quad \; \textbf{Test:} On \textcolor{red}{\{2022-10-25\}}, the historical momentum of TSLA, as measured by its cumulative logarithmic returns in the past \\
    \quad \quad \; \textcolor{red}{\{3\}} days, was \textcolor{red}{\{7.05\%\}}. \\
    \end{tcolorbox}
\end{tcolorbox}
\caption{ First section of \textsc{FinMem}'s workflow for perceiving and processing multi-sourced information from market environment.}
\label{fig:trading-prompt1}
\end{figure*}

\begin{figure*}[ht!]
\centering
\begin{tcolorbox}
    \begin{tcolorbox}[colback=white, colframe=lightorange, title=Reflect, boxsep=2pt, top=2pt, bottom=2pt]
    \small
    \raggedright
    \textbf{\textcolor{lightorange}{1. Operations:}}\\
    \textbf{Train:} - Infer the reasoning from the retrieved memories insights for the recommended trading actions and the increase or decrease of the trading stock price.\\
    \textbf{Test:} - Generate recommended trading actions. \quad \quad - Infer the reasoning for making such trading action. \\
    \textbf{Train \& Test:} - Track the memory insight IDs that contributed to the current trading decision.\\
    \textbf{\textcolor{lightorange}{2. Range:}} Interpretation of observed market indicators and top $k$ insights from each layer of the long-term memory database. \\
    \textbf{\textcolor{lightorange}{3. Prompts:}}\\
    \textbf{\textit{Train:}} Given the following information, can you explain why the financial market fluctuation from the current day to the next day behaves like this? Summarize the reason for the decision. You should provide summary information and the IDs of the information to support your summary.\\
    \textbf{\textit{Test:}} Given the information, can you make an investment decision? Just summarize the reason for the decision. Consider only the available short-term information, the mid-term information, the long-term information, and the reflection-term information.
    Consider the momentum of the historical stock price.
    Also, consider how much share of the stock the investor holds now. You should provide exactly one of the following investment decisions: buy or sell.
    You could go with the 'hold' option if necessary.
    You also need to provide the IDs of the information to support your decision.\\
    \textbf{\textcolor{lightorange}{4. Outputs:}}\\
    \textbf{Train:}\\
    - \textcolor{red}{Recommended Action Reason:}  The recommended trading action is 'Buy'. This can be attributed to positive market sentiment driven by a range of favorable factors. Notably, Tesla's strong battery supply chain, effective market expansion plans in Germany, and its ability to maintain market strength despite potential competitive threats positively influenced investor confidence, contributing to the stock price rise.\\
    - \textcolor{red}{Reflected Contributed Insights IDs:} \textcolor{red}{\{261, 278, 275, 274, 161,...\}}\\
    \textbf{Test:}\\
    - \textcolor{red}{Trading Action: \{Buy\}}\\   
    - \textcolor{red}{Recommended Action Reason:} Considering the short-term positive revenue growth, higher Model 3/Y and Model S/X deliveries, positive free cash flow, and increased cash and cash equivalents, Tesla appears to have strong financial health. The momentum being positive, it also encourages a risk-seeking strategy. The significant positive signs from news such as the imminent Q3 earnings add to the favorable view of the investment. Despite the mid-term neutral sentiment, long-term information shows only marginal drops and the stock is still looking strong overall. These, along with positive reflection-term information, form the reasoning behind the 'buy' decision...\\ 
    - \textcolor{red}{Reflected Contributed Insights IDs: \{261, 278, 275, 274, 161,...\}}\\
    \end{tcolorbox}
\end{tcolorbox}
\caption{\small Second section of \textsc{FinMem}'s workflow for generating trading action, reasoning and reflection.}
\label{fig:trading-prompt2}
\end{figure*}




\section{Trading Comparison in the Decision Making Task}
\label{trading-comparison}
\begin{figure*}[h]
    \caption{Comparison of CRs over time: FinLLaMA vs. other LLMs in TSLA trading with \textsc{FinMem}.}
\label{fig:pretraining-TSLA}
    \centering
        \includegraphics[width=0.85\linewidth]{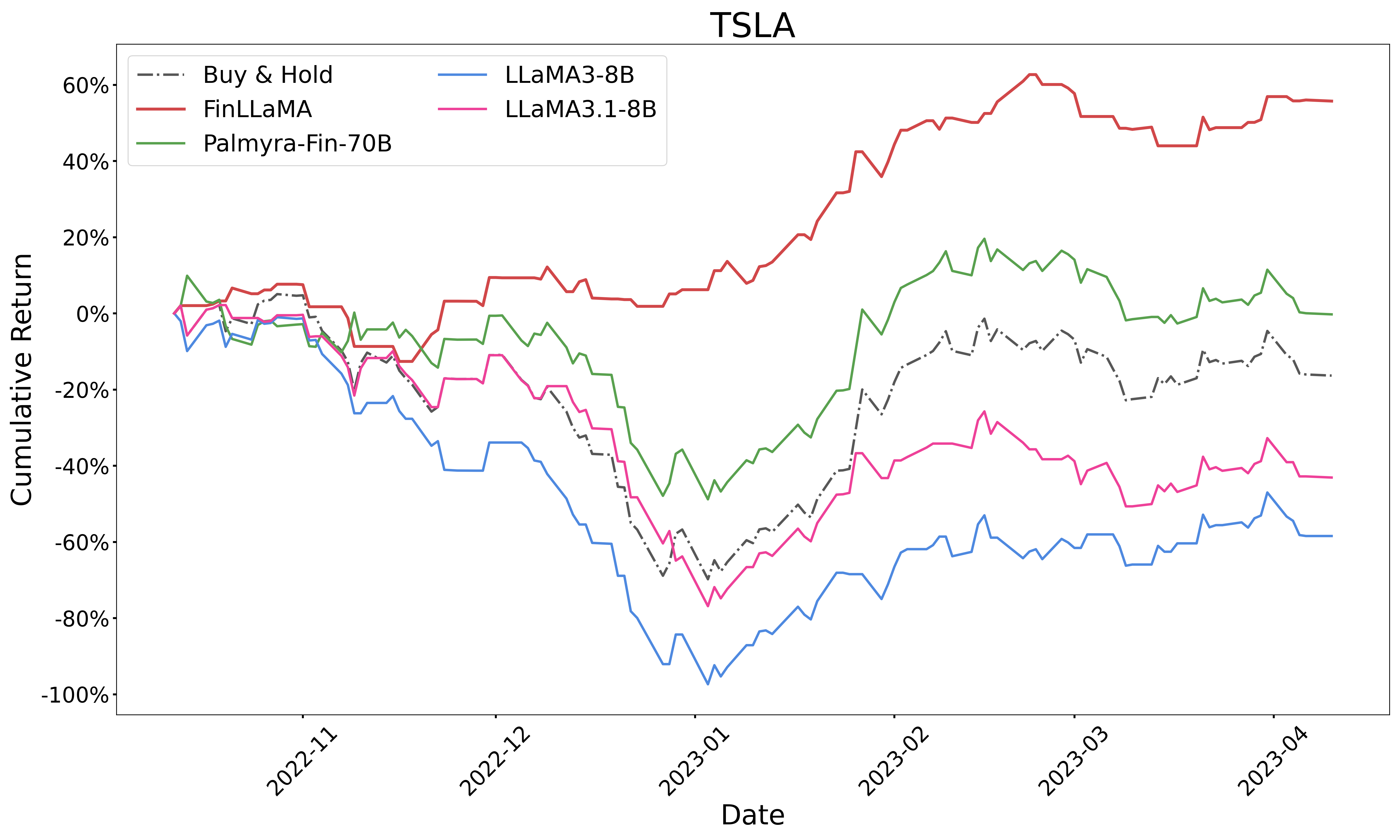}
\end{figure*}

\begin{figure*}[h]
\caption{Comparison of CRs over time: FinLLaMA vs. other LLMs in COIN trading with \textsc{FinMem}.}
\label{fig:pretraining-COIN}
    \centering
        \includegraphics[width=0.85\linewidth]{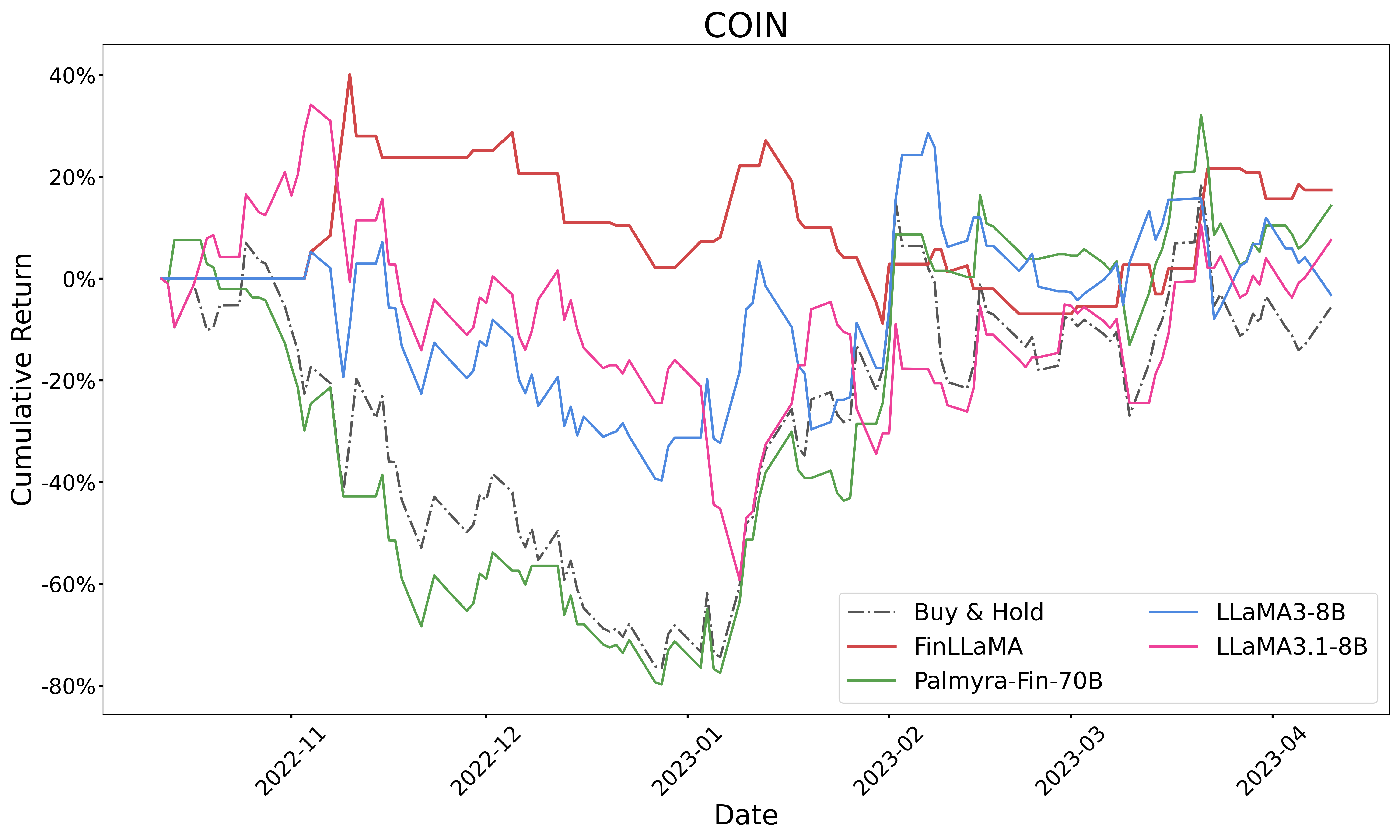}
\end{figure*}

\begin{figure*}[h]
\caption{Comparison of CRs over time: FinLLaMA vs. other LLMs in GOOG trading with \textsc{FinMem}.}
\label{fig:pretraining-GOOG}
    \centering
        \includegraphics[width=0.85\linewidth]{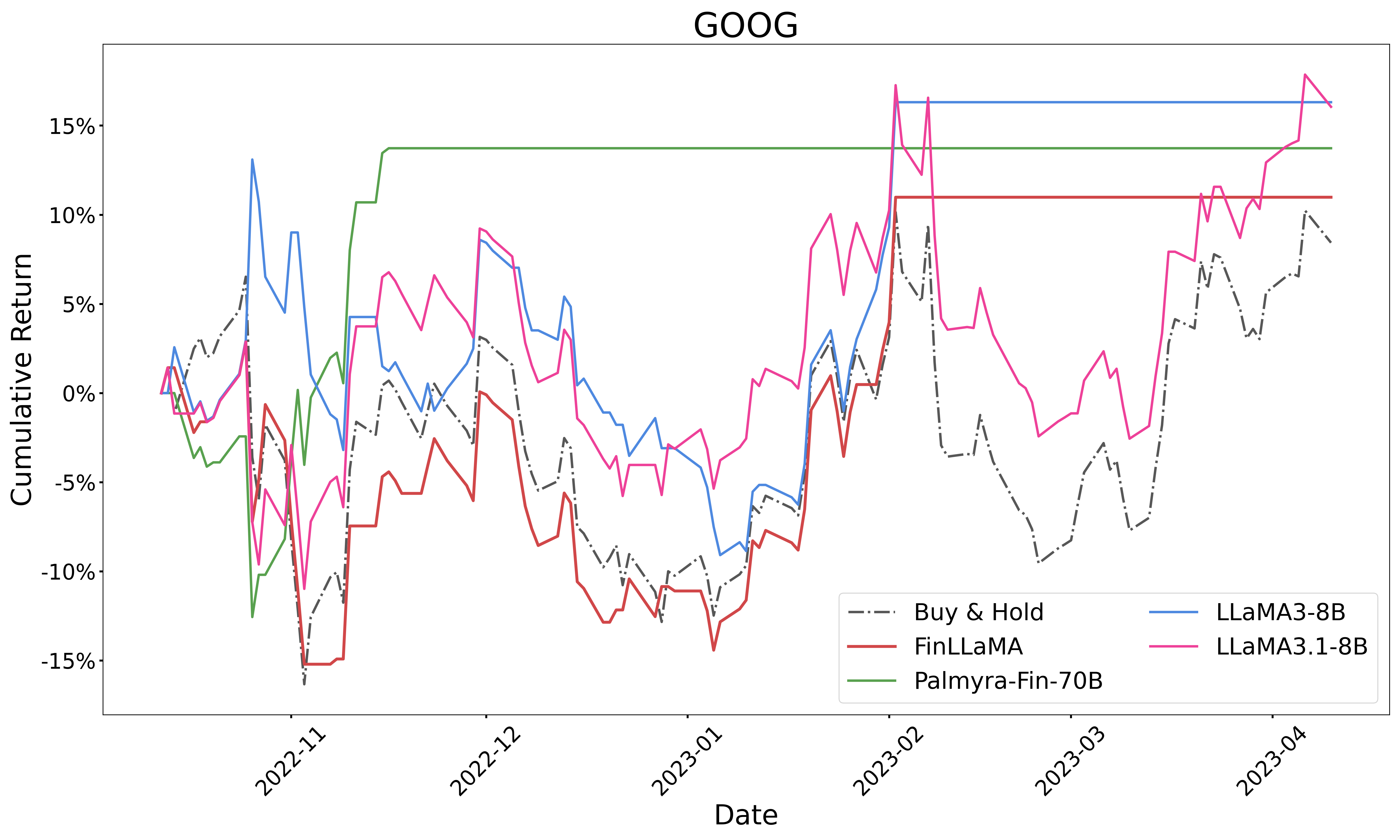}
\end{figure*}

Figure~\ref{fig:pretraining-TSLA} through~\ref{fig:pretraining-GOOG} illustrate the overall cumulative returns over time for the Decision Making task using different models. For TSLA and NIO stocks, we can see that FinLLaMA consistently outperforms other models across all time periods. On the COIN stock, FinLLaMA exhibits a more stable and consistent upward trend in cumulative returns compared to other models, particularly before February 2023. In contrast, for GOOG stock, while FinLLaMA's cumulative return is slightly lower than that of LLAMA3-8B and Palmyra-Fin-70B-32K across various time periods, it remains superior to the Buy \& Hold strategy.

\begin{figure*}[h]
\caption{Comparison of CRs over time: FinLLaMA vs. other LLMs in NIO trading with \textsc{FinMem}.}
\label{fig:pretraining-NIO}
    \centering
        \includegraphics[width=0.85\linewidth]{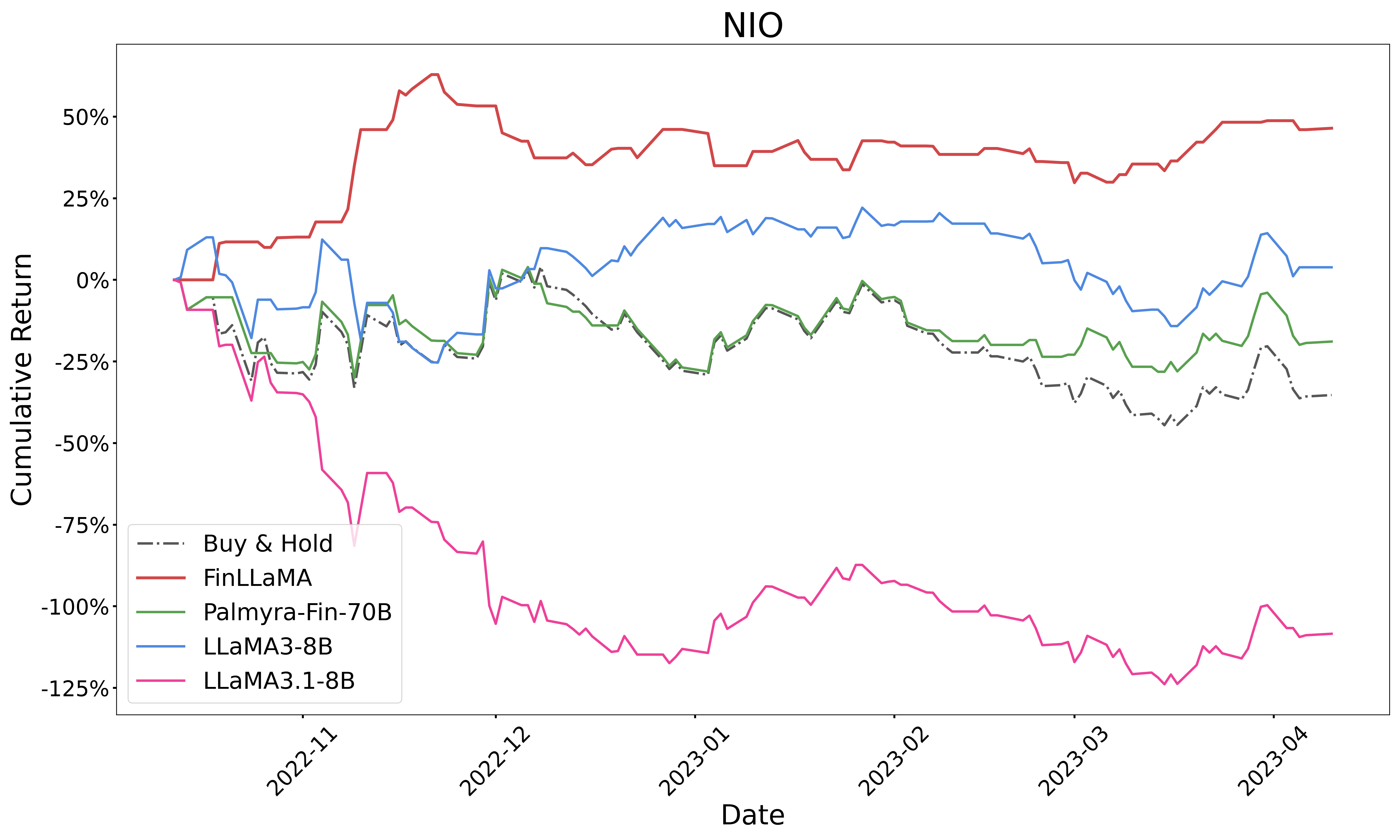}
\end{figure*}

\end{document}